\RequirePackage{fix-cm}
\documentclass[twocolumn]{svjour3}          
\smartqed  
\usepackage{url}
\usepackage{graphicx}
\usepackage{amsmath}
\usepackage{amssymb}
\usepackage{array,multirow,graphicx}
\usepackage{float}
\usepackage{makecell}
\usepackage{subfig}
\usepackage{tabu, booktabs}
\usepackage[ruled,vlined]{algorithm2e}
\usepackage{tensor}
\usepackage[T1]{fontenc}
\usepackage[font=small,labelfont=bf,tableposition=top]{caption}
\usepackage{arydshln}

\usepackage[mathscr]{euscript}
\DeclareSymbolFont{rsfs}{U}{rsfs}{m}{n}
\DeclareSymbolFontAlphabet{\mathscrsfs}{rsfs}

\DeclareCaptionLabelFormat{andtable}{#1~#2  \&  \tablename~\thetable}

\def\etal{\emph{et al}.}
\def\ie{i.e. }

\newcommand{\tref}[1]{Table~\ref{#1}}

\newcommand{\eref}[1]{Eq.~(\ref{#1})}
\newcommand{\Eref}[1]{Equation~(\ref{#1})}
\newcommand{\fref}[1]{Fig.~\ref{#1}}
\newcommand{\Fref}[1]{Figure~\ref{#1}}
\newcommand{\sref}[1]{Sec.~\ref{#1}}
\newcommand{\Sref}[1]{Section~\ref{#1}}

\renewcommand{\tabcolsep}{4pt}
\renewcommand{\arraystretch}{1.5}

\newcommand{\indep}{\rotatebox[origin=c]{90}{$\models$}}

\newcommand{\bA}{\mathbf{A}}
\newcommand{\bB}{\mathbf{B}}

\newcommand{\ba}{\mathbf{a}}

\newcommand{\bc}{\mathbf{c}}

\newcommand{\bN}{\mathbf{N}}

\newcommand{\bmu}{\boldsymbol{\mu}}

\newcommand{\bz}{\mathbf{z}}
\newcommand{\bx}{\mathbf{x}}
\newcommand{\by}{\mathbf{y}}
\newcommand{\bn}{\mathbf{n}}
\newcommand{\bm}{\mathbf{m}}

\newcommand{\as}{\stackrel{a.s.}{=}}

\DeclareMathOperator*{\argmin}{argmin} 
\graphicspath{{./image/}{./figure/}{./photos/}}

\begin{document}
\sloppy

\title{ Distance Based Image Classification:\\
A   solution  to generative classification's conundrum?
}


\author{Wen-Yan Lin         \and Siying Liu \and Bing Tian Dai \and  Hongdong Li 
}


\institute{Wen-Yan Lin \at
              Singapore Management University \\
              Tel.: +65-6826-1345\\
              \email{daniellin@smu.edu.sg}           
           \and
           Siying Liu \at
		   Institute for Infocomm Research  \\
           Tel.: +65-6408-2018\\
           \email{liusy1@i2r.a-star.edu.sg} 
           \and
            Bing Tian Dai\at
		   Singapore Management University  \\
           Tel.: +65-6828-9603\\
           \email{btdai@smu.edu.sg} 
           \and
           Hongdong Li \at
           Australia National University\\
           Tel.: +61-2-6125-7708 \\
           \email{hongdong.li@anu.edu.au} 
}

\date{Received: date / Accepted: date}

\maketitle

\begin{abstract}
Most   classifiers rely on  discriminative boundaries that separate
instances of each class from everything else. We argue that  discriminative boundaries 
are  counter-intuitive as  they define semantics by what-they-are-not;
and  should be replaced by  generative   classifiers which define  semantics  by what-they-are.  
Unfortunately, generative classifiers are 
 significantly  less accurate. This may be caused by the 
 tendency of
 generative models
      to focus    on  easy to model  semantic generative factors and   ignore   non-semantic   factors 
that are important but  difficult to model. 
We propose a new generative model  in which  semantic  factors
are accommodated by   shell theory's~\cite{lin21} hierarchical generative process
and   non-semantic   factors  by  an instance specific noise term.
We use the model to develop  a classification scheme which suppresses
the impact of noise while  
preserving    semantic cues. 
 The result is a surprisingly accurate generative classifier, that takes the form of a modified   
   nearest-neighbor algorithm; we term it   distance classification.
   Unlike  discriminative classifiers,  a distance classifier: defines semantics by
   what-they-are; is   amenable  to incremental updates; and 
     scales well with the number of classes. 
%

%

\end{abstract}

\section{Introduction}
\label{intro}

\begin{figure}[htp]
\centering
  \includegraphics[width=0.5\linewidth]{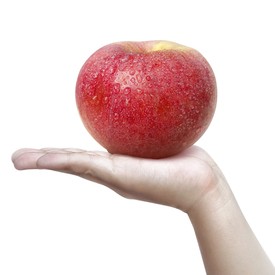}
\label{fig:1}       
\end{figure}

How do we know  apples are apples? Pondering this  question
leads to a suspicion that    discriminative classifiers, which lie at the heart of
modern machine learning, may be  conceptually flawed. 
 
In discriminative classification, each semantic class is defined  by 
a discriminative boundary which separates instances of itself from 
everything else. This  is   effective but  rather   odd, as  
semantics become defined     by what-they-are-not. Thus,  a discriminative classifier
 views an apple as a
 fruit that is  not an orange,   not a banana, and not a  pear.
 We believe many of the eccentricities that trouble computer vision are 
 caused by such, what-they-are-not semantic definitions. 
 
 Consider the task of updating a pre-trained
 discriminative classifier to incorporate  a  new   class.
 The new class will change the what-they-are-not definitions 
 of  existing classes; thus, the update requires an expensive retraining of  all  classes.  
 Failure to perform retraining will result in    catastrophic 
forgetting~\cite{chen2018lifelong}, with  old classes being progressively  forgotten as  new classes are learned. 
 Other oddities include  ill-conditioning  in the presence of  unknown semantic overlaps and 
poor scaling with number of classes,
 problems  that are unknown to  humans.  
Perhaps the solution lies in
the other major learning paradigm,
generative classification.

%

 Generative classifiers treat 
instances (images in the context of computer vision)  as  generated
outcomes and 
 labels as symbolic representations of individual generative processes~\cite{lin21}.
The mathematical linkage of instances to labels, through 
an assumed generative process,  allows  machine learning tasks to be reduced to
  mathematical problems with known solutions. This enforces a rigor, which
naturally results in semantics being defined by what-they-are.
 Unfortunately,  the  results are often disappointing.    

The generative paradigm's seminal work  is Bayesian Classification~\cite{zhang2005exploring,schutze2008introduction,rennie2003tackling},
which elegantly reduces complex learning tasks to  a
series of statistical inferences. However, despite the  guarantees of statistical   optimality,   
 Bayesian Classifiers are   often significantly less accurate than      their
 discriminative counterparts~\cite{ng2002discriminative}.

\begin{figure}%
\centering
\subfloat[A naive nearest  neighbor  classification
  is quite  accurate, affirming shell theory~\cite{lin21}.]
{\includegraphics[width=0.8\linewidth]{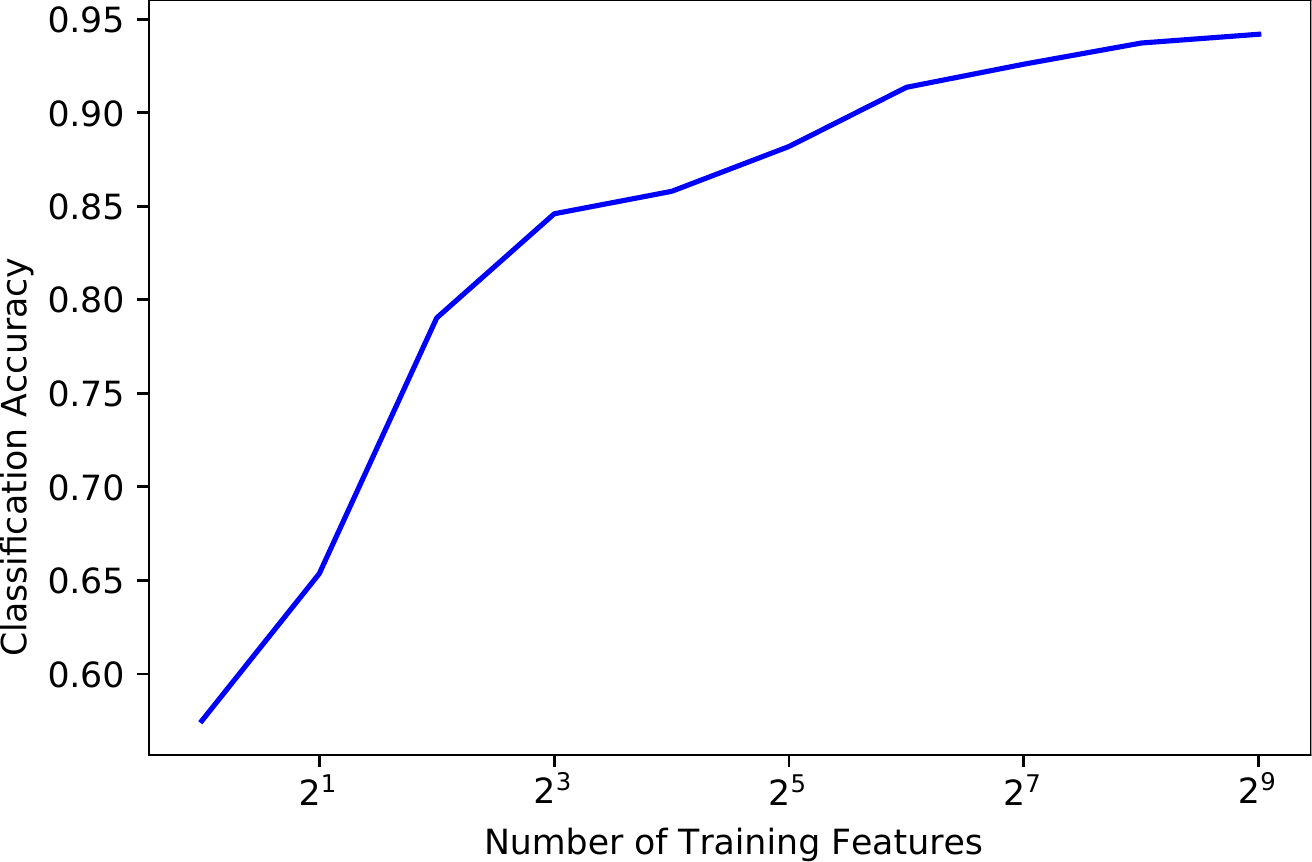}}\\
\subfloat[ Distance metrics are terrible  
			at validating class membership, contradicting shell theory.]
{\includegraphics[width=0.8\linewidth]{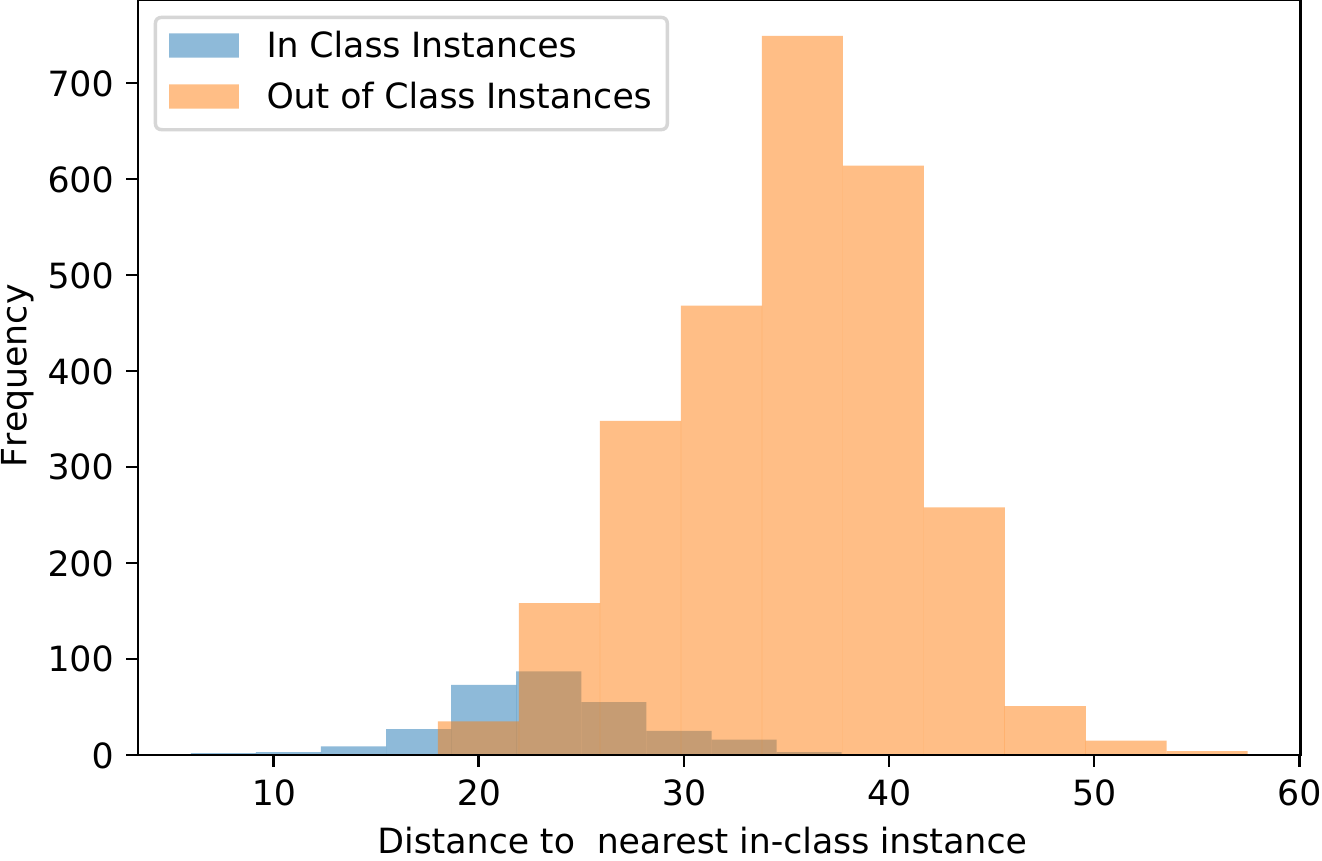}}
\caption{A  puzzling situation where distance metrics are good  for classification  but poor at validating class membership. 
Our paper suggests this    phenomenon arises from  noise induced by non-semantic generative factors
and shows how it can be corrected. }
\label{fig:distance}
\end{figure}

More recently,  shell theory~\cite{lin2018dimensionality,lin21}
employed 
the generative
framework  to show that in a high dimensional  hierarchical generative processes,
the task of classification can be reduced    to a nearest-neighbor search.
The proof is elegant. However,  
as demonstrated in~\fref{fig:distance},
naive nearest-neighbor   classifiers  display
major quirks  that  are   incompatible with shell theory's assurances of optimality.

We believe  the current problems stem from 
using unrealistic  generative models.  
Most generative models treat  instances of a semantic as i.i.d. outcomes of a common generative process;
this  implicitly assumes image appearance is solely determined 
 by  semantic factors. However,   non-semantic  factors,
  like background, lighting and object pose, also impact  image formation. 
  The challenge arises from the instance specific nature of these non-semantic  factors.
  Incorporating them into the generative 
   model,  requires  assuming   each instance has  an individualized generative process;
   this  would sever the  mathematical link between instances and labels. However,  without instance specific factors,
   the model cannot be realistic.
Our paper attempts to solve this long standing conundrum.

We propose a new generative model where 
   semantic generative factors are accommodated by the  hierarchical generative process of shell
theory~\cite{lin21}; and    non-semantic generative factors
by  an instance specific noise term.
We use this model to develop a classification scheme
that explicitly suppresses noise (non-semantic  factors) while preserving
 distance based semantic cues.
The result is a
surprisingly accurate generative classifier, that takes the form of a 
 modified nearest-neighbor algorithm;  we term it  distance classification.

 In summary, our paper provides the following theoretical   contributions:
 \begin{itemize}
 \item \textbf{New perspective on  classification:} 
 We  suggest  the popular discriminative framework is conceptually flawed.  
 \item \textbf{New approach to generative modeling:} 
 We develop a  generative model that  accommodates, often ignored,
 non-semantic  generative factors.  
 \item \textbf{Interpretable distance:} 
	We provide a  (more) realistic statistical    
	interpretation of  the distances between high dimensional entities. 
 \end{itemize}
 
From a  practical   perspective, our  contribution is the  distance
classification algorithm, which 
can be used to replace the final soft-max layer
of a (pre-trained)  neural network. 
This provides the following advantages:
\begin{itemize}

\item \textbf{Accurate:} 
The soft-max layer can be  replaced with any 
  generative classifier. However, the  distance  classifier
is uniquely accurate; thus, its use does not incur  the usual loss of accuracy.  

\item \textbf{Incrementable:} Distance classifiers 
make it trivial to  incrementally add classes to a pre-trained network.    
  Relative to prior  incremental learners,  updates with distance  classifiers  have reduced
  time complexity and greater accuracy.
 
\item \textbf{Agnostic to semantic overlaps:}  Distance classifiers
  are  agnostic to semantic
overlaps, making afore mentioned  incremental updates  more robust. 

\item \textbf{Scalable inference:}  
A soft-max layer's
  inference complexity   is $O(N)$, 
where $N$ is the number of classes. 
Distance  classifiers can employ   
approximate nearest-neighbor searches, allowing for an  inference complexity of $O(log(N))$.
This makes it possible  to accommodate a  large number of classes. 
\end{itemize}

\subsection{Related Works}
\label{sec:related}

Distance classification lies at the intersection
of a lively debate on high dimensional statistics, interpretable machine learning
and incremental learning. This section provides an overview 
of each sub-field and 
 distance classification's role in them.  
\newline

\noindent \textbf{Interpretable Distance:}
Distance is a common   classification heuristic~\cite{peterson2009k} but
 is seldom considered a formal classification constraint.
This may be because we lack a consistent mathematical framework for interpreting distances.
The problem is especially acute for distances between high dimensional entities. 

Images and image features are  high dimensional entities. According to traditional machine learning theory,
 they should be impacted by contrast-loss,
which renders high dimensional distances  meaningless~\cite{aggarwal2001surprising}.
Aggarwal \etal~\cite{aggarwal2001surprising} is often   cited to explain   algorithm failures.
However,  \fref{fig:distance} shows     nearest-neighbor classification, while
   imperfect, is not a complete failure.  
 
Shell theory~\cite{lin21} suggests  Aggarwal~\etal's  proof is flawed  because it 
assumes   all instances are independent, identically distributed outcomes of a single generative process,
thus implying the presence of only one one class. Such a generative model would   render  classification  ill-posed in both high and low dimensions; thus, it should not be used to prove that high dimensions are especially meaningless.  

Shell theory further shows that in the presence of  multiple generative processes,
  contrast-loss  makes distances
more meaningful, with the nearest-neighbor of each instance
being  its most closely related instance~\cite{lin21,lin2018dimensionality}.  
  However, shell theory is  itself inadequate. 
  \Fref{fig:distance}, shows  mutual distance is a good  classification metric
  but also a terrible metric for  class validation, simultaneously affirming
  and  contradicting 
   shell theory.   

Distance classification  extends   shell theory to incorporate  non-semantic  generative factors,
creating  a mathematical framework that explains the divergence between classification and validation scores.
The framework also allows  non-semantic factors to be suppressed,  leading to a modified  
  nearest-neighbor classifier which avoids such issues. 
\newline

\noindent \textbf{Nearest Neighbor Classification:}
Nearest neighbor classifiers assign a test instance
the  class of its nearest neighbor in a training set.
The effectiveness of this  algorithm is traditionally tied
to having sufficient training samples
to  populate the sample space densely~\cite{beyer1999nearest}.
Computer vision problems typically involve  high dimensional sample spaces,
that are too large to populate densely. Thus, it is often recommend that 
 data  be projected to a low dimensional  subspace
before performing nearest neighbor classifications~\cite{wiki_nn}.

Distance classification provides   an alternative perspective.
Like shell theory, it argues that in high dimensions,  two instances are unlikely to be coincidentally similar.
Thus, if non-semantic generative factors are suppressed,
the nearest neighbor of an instance will be   its most closely related instance. 
From this perspective,   dimensionality reduction is unnecessary;
and 
a nearest neighbor classifier with noise cancellation, can be effective with only a few training instances.
This makes the nearest neighbor algorithm
much more practical, as shown in 
\fref{fig:nn}.   
\newline

%
%
%
%

\noindent\textbf{Incremental Learning:}
Incremental learning seeks  to mimic the  human ability to add new semantic concepts 
to an  existing pool of knowledge. 
A naive solution is to fully retrain   classifiers each time a   semantic is added. However, 
retraining is   prohibitively slow.
However, as explained earlier, catastrophic forgetting~\cite{chen2018lifelong} occurs
if old classes are not retrained.

 Currently incremental solutions can divided into two main categories.  
 We term the first category weakly incremental learning. This allows retraining with past data~\cite{rebuffi2017icarl,castro2018end,wu2019large,hayes2019remind,rajasegaran2020itaml}
 but restricts it to ensure efficiency. 
 Weakly incremental learning is effective but updates incur significant computational cost. 
 The second  category is strictly incremental learning.  
 Examples include having  model  parameters  dedicated to different batches of
semantics~\cite{rajasegaran2020itaml,yoon2017lifelong,rao2019continual} or training
a series of one-class learners~\cite{van_de_Ven_2021_CVPR,chen2001one}.
Strictly  incremental learners can be updated  quickly but  are usually  less accurate.  

As distance classifiers represent classes by what-they-are, they can be incrementally
updated by  appending the appropriate class representation. 
The result is an algorithm for strict incremental learning, that is significantly more accurate than 
 weak incremental learners.
\newline

\noindent \textbf{Shell Theory~\cite{lin2018dimensionality,lin21}:}
Shell theory introduces  a statistical framework for analyzing high dimensional distance.
However, the analysis only matches empirical results for normalized data. 
Distance classification    generalizes shell theory to accommodate instance specific noise,
allowing for the interpretation of both normalized and unnormalized data.
From the perspective of distance classification,   shell theory is a special case where
  data is noise free; and normalization is one of a number of different mechanics to accommodate noise. 

As shell theory is only effective on   normalized data, it cannot be  
employed in incremental learning, as it would involve previously learned classes being re-normalized
   each time a new class is encountered.
Since distance classifiers are less dependent  on normalization,
they are more effective at   incremental learning. 
\newline

\subsection{Paper Organization}

The formulation begins  in  \sref{sec:des_main}, with  a  brief  explanation of shell theory. 
\Sref{sec:noise} and \sref{sec:noise_cancel}  introduce
noise and noise cancellation  respectively.
Finally, \sref{sec:mini}, \sref{sec:top-N} and \sref{sec:val}, discuss
  algorithm  design and provide experimental validation.

\section{Introducing Shell Theory}
\label{sec:des_main}

Statistical machine learning is typically centered on
a stochastic   model of   data generation.
These models posit that  instances
of a class share a common generative process
and are thus  concentrated in a  unique region of the 
statistical sample space.
This makes it possible to statistically
 infer an instance's  semantic class
using sample densities estimated
 from training data. 

Such density based  models are effective on  
simple machine learning tasks but cannot be 
directly applied to  complex tasks like image classification.
This is because statistical sample spaces grow exponentially
with the number of dimensions and image data is 
notably high dimensional. As a result, image 
sample spaces are often too large to  populate densely.\footnote{Image
sample spaces may be so large that most potential
images never exist. If so, it is possible that even a dataset containing all current images,
will fail to densely populate an image sample space.}

The traditional solution is to assume  data  can be captured by a low dimensional projection. This idea connects to the vast literature on low-rank methods~\cite{markopoulos2017efficient,candes2011robust,abdi2010principal,ledent2021fine} and has been applied in image recognition contexts~\cite{wu2010robust,candes2011robust}. 
 However, 
the   low-dimensional projections learned from one dataset often generalizes poorly in
the presence of  bias~\cite{linlocally,ledent2021fine};
and is not guaranteed to be appropriate  for unseen classes which 
an incremental learner will encounter.

Shell theory~\cite{lin2018dimensionality,lin21}  
is based on a different hypothesis regarding the nature of data. In shell theory,  
     the differences 
between complex semantic concepts (like ``cat'' and ``dog'') 
may be low dimensional but the concepts themselves are  high dimensional~\cite{lin21}. 
\ie  no single attribute  makes
a cat, a cat; rather,  the  concept of ``cat''    only  emerges
when multiple attributes are observed simultaneously.  
Thus, if we are to represent semantics   in terms
of what-they-are,  the problem of high dimensional
statistics needs to be addressed.  

Shell theory~\cite{lin2018dimensionality,lin21} suggests
that the  solution is to perform statistics in terms of distance,
rather than density. 
Shell theory suggests that  if each class is described   by a long list of attributes, 
 it will be highly improbable  that the description is
coincidentally satisfied by an instance from outside the
class. Shell theory then shows that this intuition can be formalized in the form of 
a generative model in which  
     high dimensional   distances act as proxies for statistical dependencies in the generative process
and vice-versa. The results is a distance based statistical framework, which  avoids the need to 
project data onto a low-dimensional subspaces;  and   accommodates unseen classes.

\subsection{Hierarchical Generative Process}
  
Apart from its focus on high dimensions, shell theory's other
major difference from classical machine learning
is its proposed,  hierarchical generative model.  

In most statistical machine learning  literature,  
  data generation is modeled using a set of stochastic generative processes. For example, a
  tiger image might be modeled as an outcome of some tiger generator.  
  This  is simple and    intuitive     but also   incomplete,
   as the tiger generator's origins    are left unexplained.

\begin{figure}[tp]
\centering
  \includegraphics[width=1\linewidth]{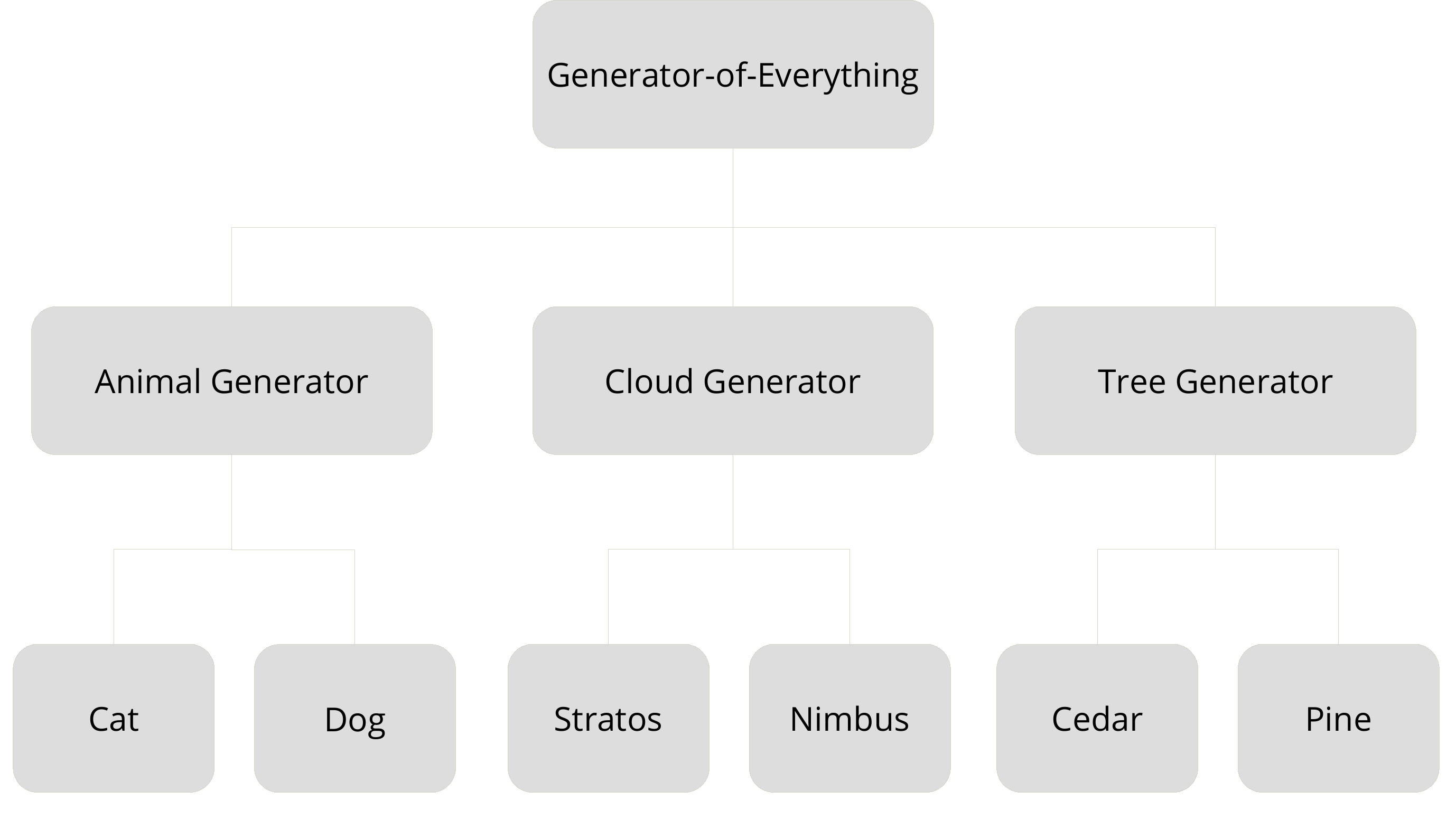}
  \caption{Shell theory assumes that  semantics are symbolic representations of a hierarchy of generative  processes.     
\label{fig:hier} }      
\end{figure}

Shell theory~\cite{lin21} addresses this issue by 
proposing a hierarchical  generative processes which 
is designed to mimic
natural generative processes. In a  
hierarchical  generative processes, 
the  tiger generator is an    outcome  of an ancestral feline generator, which
 in turn is the  outcome of an ancestral animal generator, etc. The   
 hierarchy is  rooted in a hypothesized  generator-of-everything  illustrated in \fref{fig:hier}. 

 Interestingly,  
 if  generative processes are high dimensional
 and statistically independent given their most recent common ancestor,
 shell theory predicts that  an instance's nearest-neighbor
 is almost-surely its most closely related instance. 
 If true, many classification tasks could be reduced
 to a nearest-neighbor discovery problem. 
 This section summarizes shell theory's proof, which
 we extend (in the following sections) to accommodate 
real image data. 
 
\subsection{High Dimensional Distances}
\label{sec:dist}

Following the notational conventions of statistics,
random variables are denoted with   
upper case letters like $A$.
 Random variables act as ``place-holders'' for stochastic outcomes
 that have yet to be undetermined. 
If the outcome  of  a stochastic process is known, it is termed an instance.
Instances and denoted with lower case  letters like $a$. 
The vector concatenation of two or more 
random variables is termed a random vector;  random vectors and ordinary vectors are denoted
with boldfaced types like $\bA$ and $\ba$.

Let $\bA$ and $\bB$  represent the outcomes of two \underline{independent} generative processes. As
their number of dimensions, $k$, becomes large, from~\cite{lin2018dimensionality}:
 \begin{equation}
 \label{eq:raw}
 \frac{\|\bA-\bB\|^2}{k} \as \frac{\|\bmu_\bA-\bmu_\bB\|^2 + v_\bA + v_\bB}{k},
 \end{equation}
 where  $\as$  denotes almost-surely-equal.
 $\bmu, \, v$    denote mean and total variance respectively, their subscripts indicating
 the  random variable of origin. 
 
 \Eref{eq:raw} uses  $k$ as a common denominator to ensure  individual  terms remain finite as
 $k$ tends to infinity. For readability, we  simplify  \eref{eq:raw} to:
 \begin{equation}
 \label{eq:base}
 \|\bA-\bB\|^2 \as \|\bmu_\bA-\bmu_\bB\|^2 + v_\bA + v_\bB.
 \end{equation}

 Let   $\bc$ be a constant vector,  chosen in a manner that is  \underline{independent} of $\bA$.
   We can set  $\bB$ in \eref{eq:base} to have a mean of $\bc$ and variance of  zero, thus:
 \begin{equation}
 \label{eq:base1}
 \|\bA-\bc\|^2 \as \|\bmu_\bA-\bc\|^2 +v_\bA,
 \end{equation}
\Eref{eq:base} and \eref{eq:base1} represent   the core distance relationships at the heart of our paper.

We use $\mathcal{A}$ to denote the set of generators in  a hierarchical generative process,
\begin{align*}
\centering
\mathcal{A} = \{&\mbox{generator-of-everything},\, \hdots,  \\
  &\mbox{animal generator},\, \hdots, \\
  &\mbox{cloud generator},\, \hdots, \\
  &\mbox{cat generator},\, \hdots, \\
  &\mbox{dog generator},\, \hdots, \\
 &\mbox{tree generator},\, \hdots\};
\end{align*}
 and  $\bx$ to denote an instance created by $\mathcal{A}$. 
 
In a hierarchical  generative process,   $\bx$ is simultaneously  the outcome of multiple generators.
For example, an individual tiger is simultaneously  the outcome of the tiger, feline and animal generators. 
For any $\bx$,  the  non-trivial  ancestral generators (generators with finite difference in either mean or total variance)  
are represented  by the ordered set: 
$$\mathcal{A}_\bx = \{\bA_i  \,\, |\, i \in \mathbb{Z}^+_0 \}, \quad \mathcal{A}_\bx \subset \mathcal{A},$$
where $\bA_i$ is  ancestral to $\bA_{i+1}$. 
From shell theory~\cite{lin21}, we know  that non-trivial descendant generators, almost-surely  have lower  total variance than 
their parents. Thus,  
    \begin{equation}
    \label{eq:v}
    a.s. \quad  v_{\bA_j} > v_{\bA_i}, \quad  \forall\, \bA_i, \bA_j \in \mathcal{A}_\bx,\,  i >j; 
    \end{equation}
where $a.s.$  denotes almost-surely constraints between instantiated variables.

Let $\by$ denote a different instance.
$\bx$ and $\by$ will share a most recent common ancestral generator,  $\bA_c \in \mathcal{A}_\bx$.
As the descendant generators are assumed to be independent of each other,
$\bx$ and $\by$ are independent 
outcomes of $\bA_c$. Hence,   from \eref{eq:base}:
\begin{equation}
\label{eq:neigh}
a.s. \quad \|\bx - \by\|^2 = 2v_{\bA_c}, \quad \bA_c \in \mathcal{A}_{\bx}.
\end{equation}

\Eref{eq:neigh} implies that the distance between $\bx$ and  $\by$ is
the square root of the total variance of their most  recently  shared ancestor.
\Eref{eq:v} implies that the total variance
decreases as we descend the hierarchy.
Thus,
the closer  $\bx$ is to  $\by$, the more closely related they are (share a more recent common ancestor) and vice-versa. 
If true, image classification could be reduced to a problem
of nearest-neighbor  discovery.  Unfortunately, as we  show in \sref{sec:nn_class},
in practice, the distance between  instances is corrupted by noise. To accommodate this,  
we introduce a more robust, mean based distance.

\subsection{Describing Generators by their Mean }
\label{sec:nmc}


Let  $\bx$ be an instance and 
 $\mathbb{M}$  a set of generator means, of which  at least one belongs to an  ancestral generator of $\bx$. 
 $\bmu_c$  denotes the mean  in  $\mathbb{M}$ that  is most closely related to $\bx$.
For generator-means  to act as ``descriptions'' of their respective generators,   
$\bx$ must be nearer to $\bmu_c$ than  any other mean  in $\mathbb{M}$, \ie
 \begin{equation}
 \label{eq:hypo}
 \|\bx - \bmu_c\|^2  <  \|\bx - \bmu\|^2,  \quad \forall\, \bmu \in \mathbb{M},\,  \bmu \neq \bmu_c.
\end{equation}
This section formally derives \eref{eq:hypo}.

Let $\bA_i \in \mathcal{A}_\bx$ be the generator associated with $\bmu_c$ (\ie $\bmu_{\bA_i}=\bmu_c$).
As  $\bx$  is an outcome of $\bA_i$, from \eref{eq:base1}:
\begin{equation}
\label{eq:shell_a}
a.s. \quad  \|\bx - \bmu_c\|^2 = v_{\bA_i}.
\end{equation}
Let us consider a different mean, $\bmu \in \mathbb{M},\, \bmu \neq \bmu_c$.
$\bA_j \in \mathcal{A}_\bx$, is used to denote the most recent ancestral generator of  $\bx$ and $\bmu$. From \eref{eq:base1}:
\begin{equation}
\label{eq:shell_b}
\begin{split}
a.s. \quad  \|\bx - \bmu\|^2 = v_{\bA_j} + \|\bmu_{\bA_j} - \bmu\|^2. \\
\end{split} 
\end{equation}

Observe that because  $\bmu_c$  is more closely related to $\bx$ than $\bmu$, therefore, $i>j$. Thus, from  \eref{eq:v},  
$v_{\bA_i}<  v_{\bA_j}$.
 As such,    \eref{eq:shell_a}  and  \eref{eq:shell_b} can be combined to yield:
\begin{equation}
\label{eq:key}
a.s. \quad  \|\bx - \bmu_c\|^2  <  \|\bx - \bmu\|^2,  \quad \forall\, \bmu \in \mathbb{M},\,  \bmu \neq \bmu_c.
\end{equation}
 This formally derives  \eref{eq:hypo}.
 Less formally, \eref{eq:key} proves that \textbf{the nearest generator-mean to a given feature, is   the generator-mean   most closely related to that  feature}.   

What about  the most closely related feature to a given generator-mean?
Let $\by$ be an instance that  is not descended from $\bA_i$.
If so, $\by$ must  be related to $\bmu_c$ (recall that $\bmu_c$ is the mean of $\bA_i$) by  a closest common generator, $\bA_k \in \mathcal{A}_\bx,\, i > k$.
Thus, from \eref{eq:base1} and \eref{eq:v},  the distance of $\by$  to $\bmu_c$
is almost-surely greater than that of $\bx$  to $\bmu_c$: 
\begin{equation}
\label{eq:shell}
\begin{split}
a.s. \quad  \|\by - \bmu_c\|^2  & = v_{\bA_k} + \|\bmu_{\bA_k} - \bmu_c\|^2 > v_{\bA_i} \\
\Rightarrow \quad \|\by - \bmu_c\|^2  &> \|\bx - \bmu_c\|^2 = v_{\bA_i}.
\end{split}
\end{equation}
\Eref{eq:shell}  proves that \textbf{the nearest feature to a  generator-mean, is  also the feature   most closely related to that  generator-mean}.   

\Eref{eq:key} and \eref{eq:shell} suggest the mean of a generative process 
can be used to identify 
instances of that process,  turning them into ``descriptions'' of the process.  
 
\section{Introducing Noise}
\label{sec:noise}

Shell theory makes nearest-neighbor / nearest-mean  classifiers appear  attractive.
However, the practical performance of such distance based classifiers 
is often disappoint.  We hypothesize that  this gap between theory and reality occurs because,
like  most  generative models, 
shell theory implicitly assumes
 image formation is  only impacted by  semantics.
However, in practice,  
 background,
object pose and lighting,  play  a large  role 
in determining the appearance of the final image but are not
present in the generative model.

Our solution is to group the non-semantic generative factors  into an instance specific noise term.
Thus, if $\bA$ is an  idealized generator, 
its noisy counterpart is $\bA(t)$:
\begin{equation}
\label{eq:noise_start}
\bA(t) = \bA +\bN(t), \quad \bA \in \mathcal{A}.
\end{equation}
Here, $t$ represents the instance's index and 
$\bN(t)$ the   instance specific  perturbations.

We assume that noise elements
are:  independent of one another, independent    of   the ideal   generators, 
and unbiased. More formally, denoting the  
$i^{th}$ element of  $\bN(t)$ as $\bN(t)[i]$   and  statistical independence as $\indep$, 
\begin{equation}
\label{eq:noise_zero}
\begin{split}
&E(\bN(t))= \textbf{0},& \quad \forall\, t\,;\\
&\bN(t) \; \indep \; \bA, & \quad \forall\, t,\, \bA \in \mathcal{A}\,;\\
&\bN(t) \; \indep \; \bN(t'), & \quad \forall\,   t \neq t'\,;\\
&\bN(t)[i] \;  \indep \; \bN(t)[j], & \quad \forall\,  i \neq j\,.\\
 \end{split}
\end{equation}

Let $\bc$ be a constant reference vector,  chosen in a manner which is independent of $\bA(t)$.
The distance of  $\bA(t)$ from $\bc$
can be expressed in terms of an idealized distance
with a  noise induced perturbation, using 
  \eref{eq:base1} and \eref{eq:noise_start}:
\begin{equation}
\label{eq:dream}
\begin{split}
&\|\bA(t)-\bc\|^2 \\
= &  \|\bA + \bN(t)-\bc\|^2\\
= & \|{\bA}-\bc\|^2  + 2 \times \langle \bN(t),\; \bA-\bc \rangle  +\|\bN(t)\|^2\\
= & \|{\bA}-\bc\|^2  + \Gamma (t, \bc) +  \|\bN(t)\|^2.
\end{split}
\end{equation}
When the  the number of dimensions, $k$, is  large,  \eref{eq:noise_zero} implies that
the middle gamma term tends to zero:  
\begin{equation}
\label{eq:dreamy}
\begin{split}
  \Gamma (t, \bc) 
= & 2 \times  \langle \bN(t), \;\bA(t)-\bc \rangle \\
\approx & 2 \times k \times E\left(\bN(t)[i]\right)\times E\left(\bA(t)[i]-\bc[i]\right) \\
= & 0.
\end{split}
\end{equation}
Let us study the impact of noise on idealized distances of \sref{sec:des_main}. 

\subsection{Impact of Noise on Distance}

Let $\bx(t)$ denote a noisy feature, $\bx_t$ its ideal counterpart and $\bc$
a reference vector, chosen in a manner that is independent of one 
of the generators of $\bx_t$ (recall that an instance can have multiple generators).
\Eref{eq:dream} implies that   the noisy distance, 
 $\|\bx(t) -\bc\|$ is related to the ideal  distance $\|\bx_t -\bc\|$ by:
 \begin{equation}
\label{eq:dream_}
\begin{split}
a.s. \quad \|\bx(t)- \bc\|^2 = &  \|\bx_t- \bc\|^2 +  \gamma(t, \bc) + \|\bn(t)\|^2\\
                            \approx & \|\bx_t- \bc\|^2 + \|\bn(t)\|^2;
\end{split}
\end{equation}
where $\bn(t),\,\gamma(t,\bc)$ are    instantiations of $\bN(t),\,\Gamma(t, \bc)$  
and \eref{eq:dreamy} indicates that $\gamma(t,\bc)$ is small. 

%
%

Let   $\bmu$ be  a generator-mean, \eref{eq:dream_} implies: 
\begin{equation}
\label{eq:dream_mean}
\begin{split}
a.s. \quad \|\bx(t)- \bmu\|^2 = &  \|\bx_t- \bmu\|^2 +  \gamma(t, \bmu) + \|\bn(t)\|^2\\
                            \approx & \|\bx_t- \bmu\|^2 + \|\bn(t)\|^2.
\end{split}
\end{equation}
\Eref{eq:dream_mean} is true because for   any 
given $\bx_t$ and $\bmu$, there will exist a closest common ancestral generator 
from which both can be considered independent descendants. Thus, $\bmu$ 
 fulfills the requirement for $\bc$ in \eref{eq:dream_}. 

Finally, the distance between   two noisy  features, $\bx(t), \, \bx(t')$, will be:  
\begin{equation}
\label{eq:dream_feat}
\begin{split}
a.s. \quad   & \|\bx(t)- \bx(t')\|^2 \\
= &  \|\bx_t- \bx(t')\|^2 +  \gamma(t, \bx(t')) + \|\bn(t)\|^2\\
= & \|\bx_t- \bx_{t'}\|^2 + \gamma(t, \bx(t')) + \gamma(t', \bx_t) \\
& \quad + \|\bn(t)\|^2 + \|\bn(t')\|^2\\
\approx& \|\bx_t- \bx_{t'}\|^2  + \|\bn(t)\|^2 + \|\bn(t')\|^2.
\end{split}
\end{equation}
\Eref{eq:dream_feat} comes about  because $\bx_t$ and $\bx_{t'}$ will have
a common ancestral generator.  As 
both  $\bx_t$ and $\bx_{t'}$ are   independent outcomes of their common ancestor,  we can alternately treat each instance
as $\bc$ in  \eref{eq:dream_}, yielding \eref{eq:dream_feat}.

\Eref{eq:dream_}, \eref{eq:dream_mean} and \eref{eq:dream_feat}  show 
that noise induced perturbations can be decomposed into  two terms:
\begin{itemize}
\item \textit{Gamma perturbations}: $\gamma(t, \_)$, that depend on both the instance in question, $t$,
and the reference vector. $\gamma(t, \_)$ tends to be small;
\item \textit{Noise magnitude perturbations}: $\|\bn(t)\|^2$   are  constant  for any given instance,  $t$.
\end{itemize}
Of the two terms, it is most important  to cancel out the dominant,   noise magnitude perturbation. 
Fortunately, it is  constant for any given instance, $t$, making 
it possible to remove the noise induced perturbations   through carefully selected distance comparisons.

\subsection{Noise Invariant Nearest Mean Classifiers}
\label{sec:noise_ncmc}

Before explaining our solution for noise cancellation, we illustrate the implicit 
noise cancellation   embedded in  a 
 nearest mean classifier. 
%
%

Let  $\mathbb{M}$ be a  set of 
generator-means\footnote{A generator-mean can be estimated by averaging  a large number
of its generated instances. As noise has a mean of zero, this estimate is unbiased.}.

 $\bx(t)$ is a noisy test feature. From 
\eref{eq:dream_mean}, the distance of $\bx(t)$ to a generator-means is:
\begin{equation}
\label{eq:d_mean}
\begin{split}
a.s. \quad   & \|\bx(t)- \bmu\|^2 \\
= &\|\bx_t- \bmu\|^2 +  \gamma(t, \bmu) + \|\bn(t)\|^2\\
\approx & \|\bx_t- \bmu\|^2 + \|\bn(t)\|^2, \quad \quad \quad \quad \forall \bmu\in \mathbb{M}.
\end{split}
\end{equation}

Thus, if $\bmu_c$ is the generator-mean in  $\mathbb{M}$  most closely related  to $\bx_t$,
from    \eref{eq:key}   and  \eref{eq:d_mean}: 
\begin{equation}
\label{eq:noise_ncmc}
\begin{split}
a.s. \quad   &  \|\bx(t)- \bmu\|^2 - \|\bx(t)- \bmu_c\|^2 \\
= &\|\bx_t- \bmu\|^2 - \|\bx_t- \bmu_c\|^2 +  \gamma(t, \bmu) - \gamma(t, \bmu_c)\\
 &+  \|\bn(t)\|^2 - \|\bn(t)\|^2\\
= &\|\bx_t- \bmu\|^2 - \|\bx_t- \bmu_c\|^2 +  \gamma(t, \bmu) - \gamma(t, \bmu_c)\\
\approx & \|\bx_t- \bmu\|^2 - \|\bx_t- \bmu_c\|^2 >0,\\
 &\quad \quad \quad \quad \quad  \quad  \quad  \quad \quad \quad \quad  \forall   \bmu \in \mathbb{M},\, \bmu \neq \bmu_c.
\end{split}
\end{equation}
\Eref{eq:noise_ncmc} shows that  
the nearest generator-mean to $\bx(t)$ is also the generator-mean  most closely related  to $\bx(t)$,
a property that is happily noise  invariant.

The noise invariance in  \eref{eq:noise_ncmc} arises from the act of 
  comparing $\bx(t)$ to a number of different means.
The dominant, noise magnitude perturbation is constant for a given $t$,
and is cancelled by the  comparisons.

Noise cancellation becomes even stronger if  gamma perturbations  
vary smoothly with $\bmu$. If so, the already weak gamma perturbations 
would be  further weakened by the partial self cancellation in
\eref{eq:noise_ncmc}. The self cancellation effect is  strongest
for the  most important comparisons, those that take place between similar means.

Unfortunately,   noise cancellation is not to  be taken for granted. Let us demonstrate it by   reversing the classification problem.
Instead of seeking the generator-mean  most closely related to a feature,
we  seek the feature most closely related to  a  generator-mean. 

Let us assume $\bmu_c$ is more closely related to $\bx_t$ than $\bx_{t'}$. 
From \eref{eq:shell}, 
\begin{equation}
\label{eq:repeat_dd}
a.s. \quad \|\bx_{t'}- \bmu_c\|^2 - \|\bx_t- \bmu_c\|^2 > 0.
\end{equation}
This allows  the distance of an instance  from a given generator-mean, to act as an indicator 
of the closeness of their relationship.  In
 the presence of noise, \eref{eq:repeat_dd} becomes:
\begin{equation}
\label{eq:dd_wrong}
\begin{split}
a.s. \quad   & \|\bx(t')- \bmu_c\|^2 - \|\bx(t)- \bmu_c\|^2   \\
= &\|\bx_{t'}- \bmu_c\|^2 +  \gamma(t', \bmu_c) + \|\bn(t')\|^2\\
&\quad -\|\bx_{t}- \bmu_c\|^2 -  \gamma(t, \bmu_c) - \|\bn(t)\|^2\\
\approx  &  \|\bx_{t'}- \bmu_c\|^2  -\|\bx_{t}- \bmu_c\|^2 + \|\bn(t')\|^2 - \|\bn(t)\|^2.
\end{split}
\end{equation}
This may be greater than,  less than, or equal to zero, depending on the noise magnitude terms, 
$ \|\bn(t')\|^2 - \|\bn(t)\|^2$, which are not  small. 

Unlike the ideal case  in \eref{eq:repeat_dd}, 
in the presence of  noise, \eref{eq:dd_wrong} shows that the 
feature nearest to a generator-mean, is not necessarily  the feature  most closely
related to the  generator-mean. Thus, the  validity of  assigning a feature
to a generator (label validation), cannot be measured by the  raw distance between the  feature and
the generator-mean.  

\begin{figure}
\centering
\subfloat[Raw distance to class-means can be used
for classification measure but not  for 
validating class membership.   \Sref{sec:noise_ncmc} 
explains the phenomenon
 analytically.  \Sref{sec:noise_cancel} corrects the problem through noise
cancellation. ]{
\label{subfig:correct}
\begin{tabular}{l|c|cc}
\hline
& \multirow{2}{*}{\makecell[c]{Raw \\distances}}  & \multicolumn{2}{c}{\shortstack{Noise cancellation \\with reference vector}} \\
&  & (random vector)&    (mean)  \\
\hline
\makecell[c]{nearest-mean\\classification\\(accuracy)} & 0.945 & 0.945 & 0.945\\
\hline
\makecell[c]{label validation\\(auroc)}   & 0.874 & 0.965 & 0.992 \\
\hline
\end{tabular}}
\\
\subfloat[Histogram of distance from airplane's class mean, before and after noise cancellation. Noise 
cancelled   distances  can be a validation measure for class membership.]{
\label{subfig:notwhitelight}
\centering
\begin{tabular}{cc}
\includegraphics[width=0.45\linewidth]{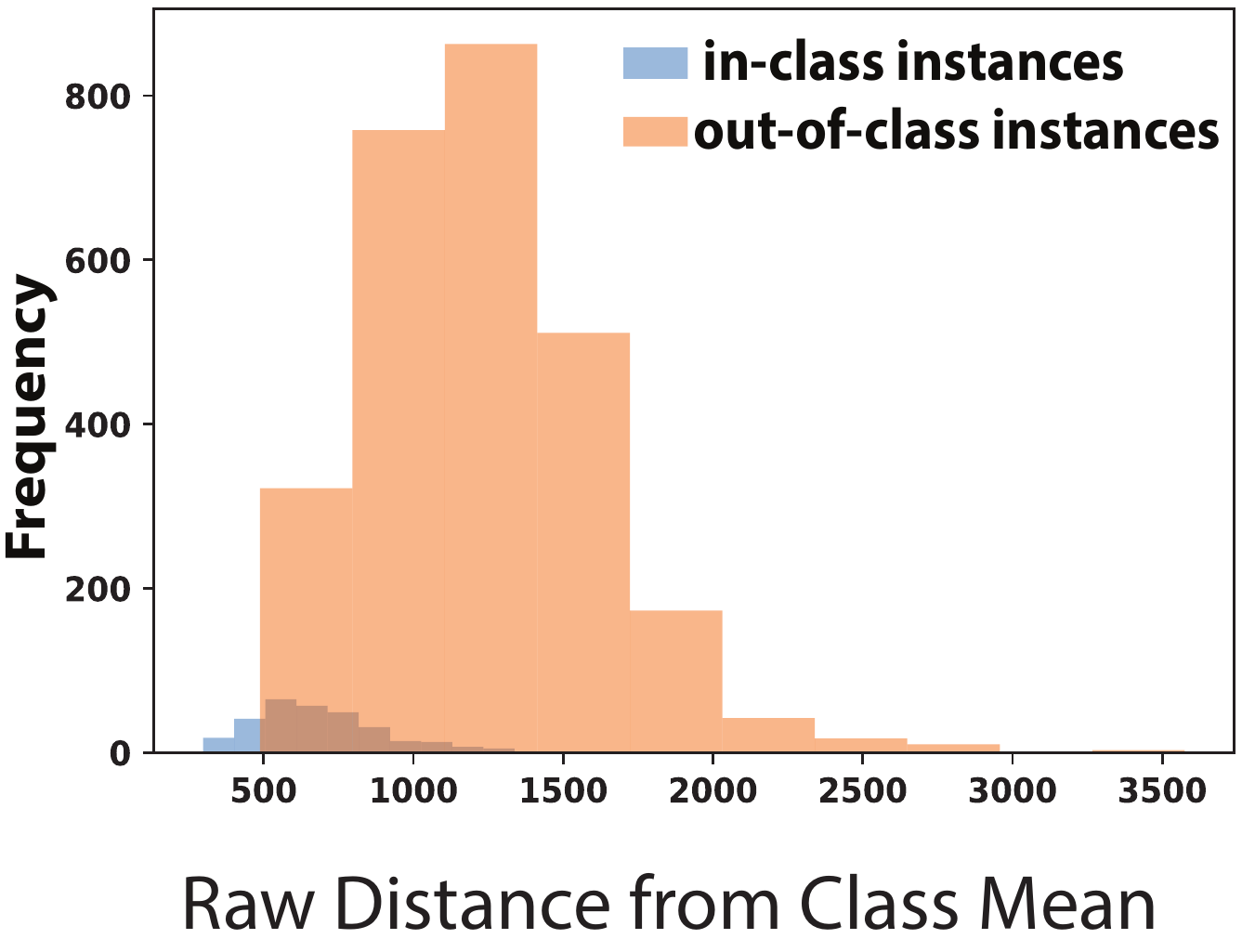} & \includegraphics[width=0.45\linewidth]{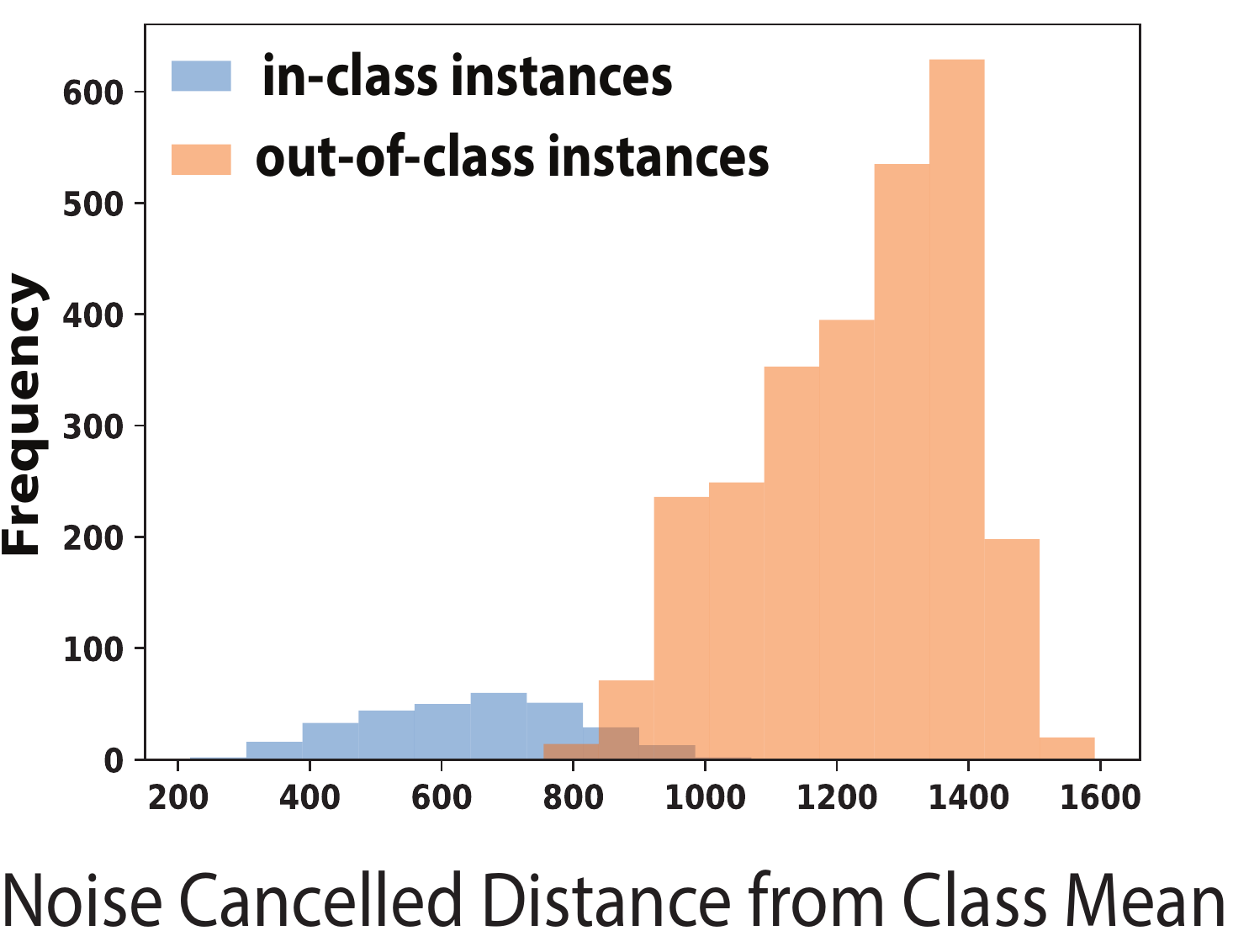}
\end{tabular}
 }
\caption{Experiments on STL-10~\cite{coates2011analysis}.  Images are represented with ResNet-50 features~\cite{he2016deep}. \label{fig:strange}}
\end{figure}

In summary, the raw distance of a feature to a generator-mean
is a good classification measure    but a terrible validation measure.
This is empirically  verified  in  \fref{fig:strange}.  
To our knowledge, we are the first paper to report this peculiar divergence in classification
and validation performance.  
%
%

\section{Noise Cancellation for Label   Validation}
\label{sec:noise_cancel}

If the distance of a feature to a generator-mean is to be directly interpretable,
the ``signal-to-noise-ratio'' needs to be modified. This section proposes
three different solutions. 
\Sref{sec:ref_cancel} uses a reference vector based noise cancellation to explicitly estimate
noise cancelled distances.
\Sref{sec:ratio_cancel} shows that certain ratios are both intuitive  and noise cancelling. 
\Sref{sec:norm} uses normalization to enhance underlying distance based constraints, creating a pre-processor
for downstream algorithms.

\subsection{Reference Vector  Based Noise Cancellation}
\label{sec:ref_cancel}

Reference vector based noise cancellation, uses  a comparison
to an appropriately chosen reference vector for noise cancellation. 

Let   $\mathbf{E} \in \mathcal{A}$ be  the generator-of-everything
and $\bz$  an independently  selected reference vector;
from \eref{eq:base1}:
\begin{equation}
\label{eq:z}
\|\mathbf{E}-\bz\|^2 \as \|\bmu_\mathbf{E}-\bz\|^2 +v_\mathbf{E} = c_\bz,
\end{equation}
where $c_\bz = \|\bmu_\mathbf{E}-\bz\|^2 +v_\mathbf{E}$ is a constant. 
If $\bx_t$ is an  ideal  feature, it is necessarily     an outcome of the  generator-of-everything;
thus, \eref{eq:z} implies     the distance of $\bx_t$  from  $\bz$ is:
\begin{equation}
a.s. \quad \|\bx_t - \bz\|^2 = c_\bz, \quad  \forall t. 
\end{equation}
$\bx(t)$ is the noisy counterpart of $\bx_t$,
thus from \eref{eq:dream_}: 
\begin{equation}
\label{eq:noise_basic}
\begin{split}
a.s. \quad \|\bx(t)-\bz\|^2  
= & \|\bx_t - \bz\|^2  +  \gamma(t, \bz) + \|\bn(t)\|^2\\
= & c_\bz  +  \gamma(t, \bz) + \|\bn(t)\|^2\\
\approx & c_\bz + \|\bn(t)\|^2, \quad \quad \quad \quad   \quad \quad \forall t.
\end{split}
\end{equation}
From \eref{eq:dream_mean}, the distance of $\bx(t)$ from a generator-mean is:
\begin{equation}
\label{eq:noise_mean}
\begin{split}
a.s. \quad  & \|\bx(t)-\bmu\|^2  \\
= & \|\bx_t-\bmu\|^2  +  \gamma(t, \bmu) + \|\bn(t)\|^2\\
\approx & \|\bx_t-\bmu\|^2 + \|\bn(t)\|^2,  \quad \quad \quad \quad \forall t, \, \forall \bmu \in \mathbb{M},
\end{split}
\end{equation}
where $\mathbb{M}$ is a set of generator-means. 

By subtracting \eref{eq:noise_basic}, from \eref{eq:noise_mean} we can eliminate
 the primary noise perturbation, $\|\bn(t)\|^2$: 
\begin{equation}
\label{eq:noise_cancel}
\begin{split}
 & d^2(\bx(t), \bmu, \bz) \\
=& \|\bx(t)- \bmu\|^2 -  \|\bx(t)- \bz\|^2 \\
=& \|\bx_t- \bmu\|^2 +  \gamma(t, \bmu) - \gamma(t, \bz) - c_\bz   \\
\approx &   \|\bx_t- \bmu\|^2 - c_\bz, \quad \quad \quad \quad \quad  \quad \quad \quad \forall t, \, \forall \bmu \in \mathbb{M}.
\end{split}
\end{equation}
Thus,  the noise cancelled distance, $ d^2(\bx(t), \bmu, \bz)$,  is a noise invariant estimate of $\bx(t)$'s  distance
to $\bmu$, with a constant   offset  of -$c_\bz$.

From \eref{eq:shell} and  \eref{eq:noise_cancel},  we know  that  if $\bmu_c \in \mathbb{M}$ is more closely related to  $\bx_t$    than 
 $\bx_{t'}$:
\begin{equation}
\label{eq:dd_right}
\begin{split}
  \quad \|\bx_{t'}- \bmu_c\|^2 & > \|\bx_t- \bmu_c\|^2, \\
\Rightarrow   \quad d^2(\bx(t'), \bmu_c, \bz) & \gtrapprox d^2(\bx(t), \bmu_c, \bz).\\
\end{split}
\end{equation}
\Eref{eq:dd_right} implies that  the noise
cancelled distance, $ d^2(\bx(t), \bmu_c, \bz)$, is an indicator for  how closely   related
$\bx(t)$ is to  $\bmu_c$. 
\newline

\noindent\textbf{Common Mean as Reference Vector:}
Let us assume that the generator-means in $\mathbb{M}$ are stochastic outcome of a most recent common 
ancestor 
$\mathbf{E}$\footnote{Unlike in  \eref{eq:z},  in this context,  $\mathbf{E}$  need not be the generator-of-everything. 
For example, if the dataset consists  of different cat species,  $\mathbf{E}$ would be the feline generator.  
}.
$\bm$  is used to denote the mean of $\mathbf{E}$.
Noise cancellation is enhanced  if   $\bm$ is the reference vector in \eref{eq:noise_cancel} (\ie $\bz=\bm$).
  The explanation is as follows.

If    gamma perturbations vary  smoothly with choice of  reference vector, $\bz$,
 noise cancellation in \eref{eq:noise_cancel} will be most effective when  $\bz$ is
as close as possible to individual means in $\mathbb{M}$,  while being chosen in a manner that is independent of $\mathbf{E}$.
$\bm$  satisfies  both properties, as explained below.  

 First, the mean of $\mathbb{M}$ is the vector which minimizes   the  average distance to each element in
   $ \mathbb{M}$. From the towering property of conditional expectation,
   the expected mean of $\mathbb{M}$ is the mean
   of the overall generator, $\mathbf{E}$, which is  $\bm$. Second,
   $\bm$ is generated during the creation of generator  $\mathbf{E}$.
   As $\bm$  is created prior to any instances of $\mathbf{E}$, its creation
   is independent of $\mathbf{E}$.
    
From \eref{eq:base1}, we know  all  instances descended from   $\mathbf{E}$ will be at a constant distance
from $\bm$.  $c_\bm$ is used to denote the square of this  distance. Thus,
\begin{equation}
a.s. \quad \|\bx_t-\bm\|^2  = c_\bm, \quad \forall t \in \mathcal{T},
\end{equation}
where the set  $\mathcal{T}$ contains all indices of features generated by $\mathbf{E}$. 
Similar to    \eref{eq:noise_cancel}, the noise cancelled distance with $\bm$ as reference vector is:
\begin{equation}
\label{eq:noise_cancel1}
\begin{split}
 & d^2(\bx(t), \bmu, \bm) \\
 =& \|\bx(t)- \bmu\|^2 -  \|\bx(t)- \bm\|^2 \\
=& \|\bx_t- \bmu\|^2 +  \gamma(t, \bmu) - \gamma(t, \bm) - c_\bm  \\
\approx & \|\bx_t- \bmu\|^2  - c_\bm, \quad \quad \quad \quad \quad  \quad \forall t \in \mathcal{T}, \, \forall \bmu \in \mathbb{M}.
 \end{split} 
 \end{equation}
The approximation in \eref{eq:noise_cancel1} is better than that in 
\eref{eq:noise_cancel}, because it enhances the partial self cancellation   of  gamma perturbations, 
 $ \gamma(t, \bmu)- \gamma(t, \bm)$. The impact 
 on validation accuracy  is illustrated in \fref{fig:strange}.

\subsection{Ratio Based Noise Cancellation}
\label{sec:ratio_cancel}

Distance ratios are often more interpretable than 
raw distances.
A famous example  is  SIFT's ratio 
test~\cite{lowe1999object}, where the best feature match is validated
by its ratio to the second best feature match.

Interestingly, the ratio of  $\bx(t)$'s distance to means 
$\bmu_i$ and  $\bmu_j$, is noise invariant  when
the distances $\|\bx_t - \bmu_i\|^2, \, \|\bx_t - \bmu_j\|^2$ are similar.  
This allows  classification thresholds to be defined in terms of 
appropriately chosen distance ratios.


Let $\bx(t)$ be  a feature  and $\bx_t$ its ideal counterpart;  $\bmu_i, \bmu_j \in \mathbb{M}$ be a pair of means  whose distances
to  $\bx(t)$ are similar; 
\ie
\begin{equation*}\Delta_{ij} = \|\bx_t - \bmu_i\|^2 - \|\bx_t - \bmu_j\|^2,
\end{equation*}
is small relative to either  $\|\bx_t - \bmu_i\|^2$ or $\|\bx_t - \bmu_j\|^2$.   Noise perturbations, 
\begin{equation*}
\gamma(t, \bmu_i) + \|\bn(t)\|^2, \quad \quad  \gamma(t, \bmu_j) + \|\bn(t)\|^2,
\end{equation*}
are also assumed to be small relative to  either $\|\bx_t - \bmu_i\|^2$ or $\|\bx_t - \bmu_j\|^2$.

The ratio of $\bx(t)$'s distance to  $\bmu_i,\, \bmu_j$ is: 
\begin{equation}
\label{eq:ratio}
\begin{split}
&\frac{\|\bx(t)  - \bmu_i \|^2}{\|\bx(t)  - \bmu_j \|^2}  \\
= & \frac{\|\bx_t  - \bmu_i \|^2 + \gamma(t, \bmu_i) + \|\bn(t)\|^2}{\|\bx_t - \bmu_j\|^2 + \gamma(t, \bmu_j) + \|\bn(t)\|^2} \\
\approx & \frac{\|\bx_t  - \bmu_i \|^2}{\|\bx_t  - \bmu_j \|^2}
  \times \left(1 + \frac{\gamma(t, \bmu_i) + \|\bn(t)\|^2}{\|\bx_t - \bmu_i\|^2 }\right) \\
& \quad \times  \left(1 - \frac{\gamma(t, \bmu_j) + \|\bn(t)\|^2}{\|\bx_t - \bmu_j\|^2 }\right)\\
\approx & \frac{\|\bx_t  - \bmu_i \|^2}{\|\bx_t  - \bmu_j \|^2} \\
& \quad \times \left(1 + \frac{\gamma(t, \bmu_i) + \|\bn(t)\|^2}{\|\bx_t - \bmu_j\|^2 + \Delta_{ij}}
- \frac{\gamma(t, \bmu_j) + \|\bn(t)\|^2}{\|\bx_t - \bmu_j\|^2} \right)\\
\approx & \frac{\|\bx_t  - \bmu_i \|^2}{\|\bx_t  - \bmu_j \|^2} \left(1 + \frac{\gamma(t, \bmu_i)-\gamma(t, \bmu_j )}{\|\bx_t - \bmu_j\|^2}\right).
\end{split}
\end{equation}
Observe that the  first order terms of \eref{eq:ratio} contain only  gamma terms,
that are both   small and  self cancelling. Thus, the    ratio
\begin{equation}
\label{eq:ratio_final}
\frac{\|\bx(t)  - \bmu_i \|^2}{\|\bx(t)  - \bmu_j \|^2} \approx  \frac{\|\bx_t  - \bmu_i \|^2}{\|\bx_t  - \bmu_j \|^2},
\end{equation}
is   noise invariant.

The strength of  using distance ratios is they can be adapted for 
fine-grained comparison. Their  drawback is that noise cancellation 
is only valid for   similar distance; this   restricts 
   ratio usage   to  local comparisons.

\subsection{Shell Normalization}
\label{sec:norm}

Shell theory~\cite{lin21} suggests a normalization
step is required to  enhance  distance based classification constraints.
The procedure is as follows:  
 Let $\bm$ be the  mean of the most recent  ancestral generator of a set of  features. 
 $\bm$ is subtracted from every feature and the resultant vector is then divided by
 its magnitude, such that it becomes  a unit-vector:
\begin{equation} 
\label{eq:norm}
\hat{\bx} = \frac{\bx-\mathbf{m}}{\|\bx-\mathbf{m}\|},
\end{equation}
where $\hat{\bx}$ denotes the normalized version of $\bx$. 

From the current paper's perspective,
 normalization is a form of  noise cancellation,
 with the associated rescaling reducing the impact of noise while 
 also  enhancing   ``shell distinctiveness'', as explained in shell theory~\cite{lin21}).

Normalization's advantage   is it creates ``better''    features. Thus, 
normalization can   
boost the accuracy of downstream algorithms, like clustering,   without 
requiring  explicit modification of the algorithm.  
However, normalization's feature transformation    can also be a disadvantage. For example,
in  incremental learning, the desired normalization 
changes as  classes are added. Ensuring the normalization is uptodate
would require re-normalizing instance of previously learned classes, as new classes
are introduced,  significantly slowing
the incremental update. In this scenario,
 ratio based noise cancellation from \sref{sec:ratio_cancel} would  be more appropriate.

\section{Training a Distance Classifier}
\label{sec:mini}

The previous sections have focused on  assigning each feature
to its most closely related   generative process. However,  classification
tasks are  formulated in terms of  labels, not  generative processes. 
Thus,  a mapping from labels to generative processes, needs to be established.

The naive solution is  to  assume each  label corresponds to a unique generative process
and represent each 
semantic  by the mean of its features. Empirical results in \fref{fig:strange} and shell theory~\cite{lin21} show
  this is   effective.  However, it may not be ideal.  

\begin{figure}[t!]
\centering
\includegraphics[width=1\linewidth]{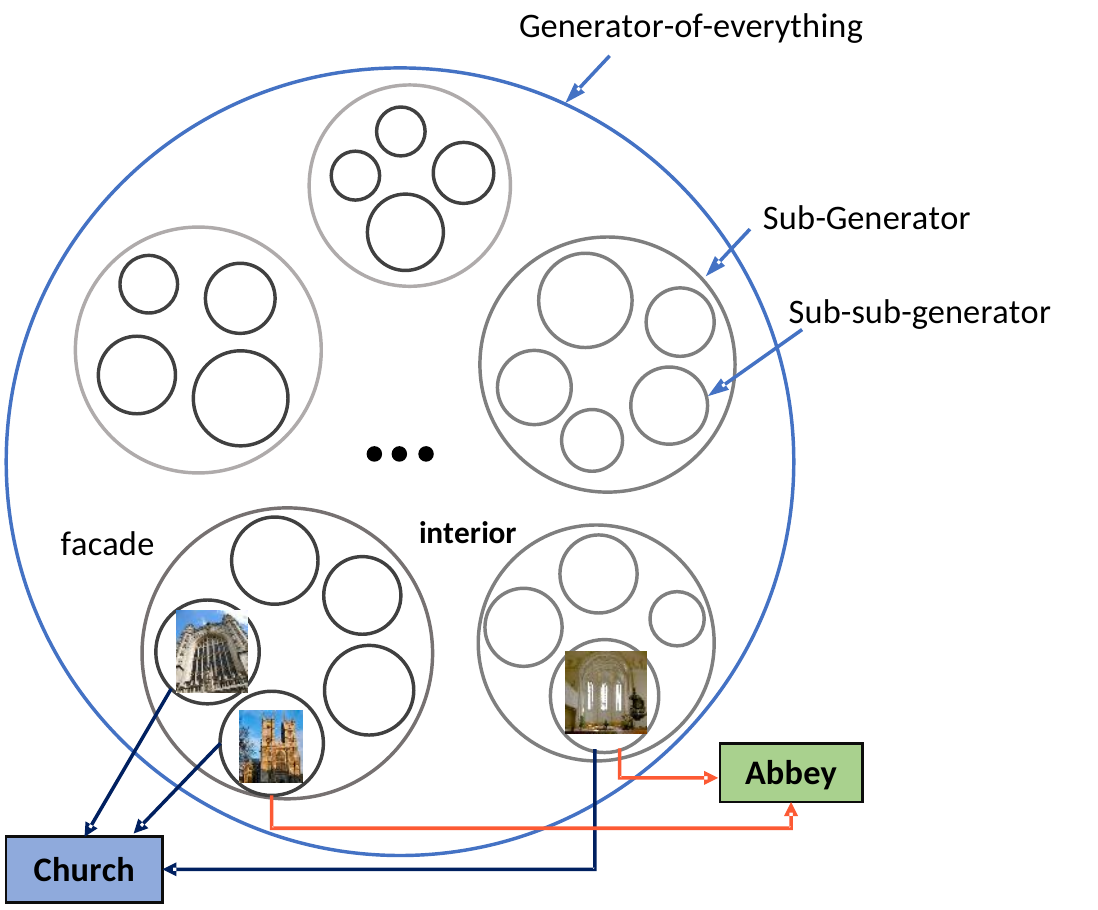}\\
\caption{
Labels are symbolic representations of   an ensemble of generative  processes, not all of which 
are closely related. Each label should be   represented a corresponding  ensemble of means,
which can be estimated through K-Means clustering~\cite{kanungo2002efficient}.    \label{fig:mini}}
\end{figure}

Let us consider the label, ``church''. 
This label can be applied to both the  interior or facade of a christian religious  building. 
 However,  church interiors are likely to   be more closely related
to the interiors of other buildings, than they are to church facades.
Likewise,  church facades are likely to be  more closely related to the  facades of other buildings, than to the church interiors. This
suggests  labels are symbolic representations of  an
ensemble of individual generative processes, not all of which are closely related. This is  illustrated in \fref{fig:mini}. 

This  perspective simultaneously   explains the complex semantic overlaps common in our language;
and allows that the semantic mean can be   a  good identifier for instances of a semantic. 
After all,   an instance that is unusually close to one of the ensemble of generative means,  is  likely to  
also be unusually close to the overall semantic mean. 

Thus, although  labels can be represented with a single mean,
it is  better to represent them  with   an ensemble of means, that correspond 
as closely as possible to the
individual generator-means. In practice,  these means 
can be estimated through  K-Means clustering~\cite{kanungo2002efficient} of a label's features:
\begin{itemize}
\item Prior to clustering, all features of a class are normalized with respect to the
 semantic  mean, providing  implicit noise cancellation (\sref{sec:norm}).
\item  The number of clusters is set at  $K=30$; this (large) number should suffice  ensure each   
 unrelated process is represented by an individual  cluster.
\item To accommodate K-means clustering's  randomness,  
the clustering procedure is repeated $r$ = $10$ times.
\item Generator-means are estimated by taking the mean of the \underline{un-normalized} features in each cluster.
The set of estimated generator-means, is used as a representation of  the semantic.
\end{itemize}
 The process is summarized in Algorithm \ref{algo:clustering} and  represents a Distance Classifier's
  training step. 
 The semantic label of a test feature is inferred from the test feature's distance to the stored generator-means. 
  Two different forms of inference 
are discussed, Distance Top-N  in \sref{sec:top-N} and Distance (VS), (VW) in  \sref{sec:val}. 

\subsection{Design Choices}
\label{sec:ablation}

Readers may have noticed  the proposed classifier uses a very large
 representation size. This design choice was made because employing multiple
 cluster-means improves 
accuracy but does not detract from  scalability. 
We explain the design choice below. 

\begin{figure*}[htp]
\centering
\begin{tabular}{cc}
MNIST & STL-10 \\
\includegraphics[width=0.4\linewidth]{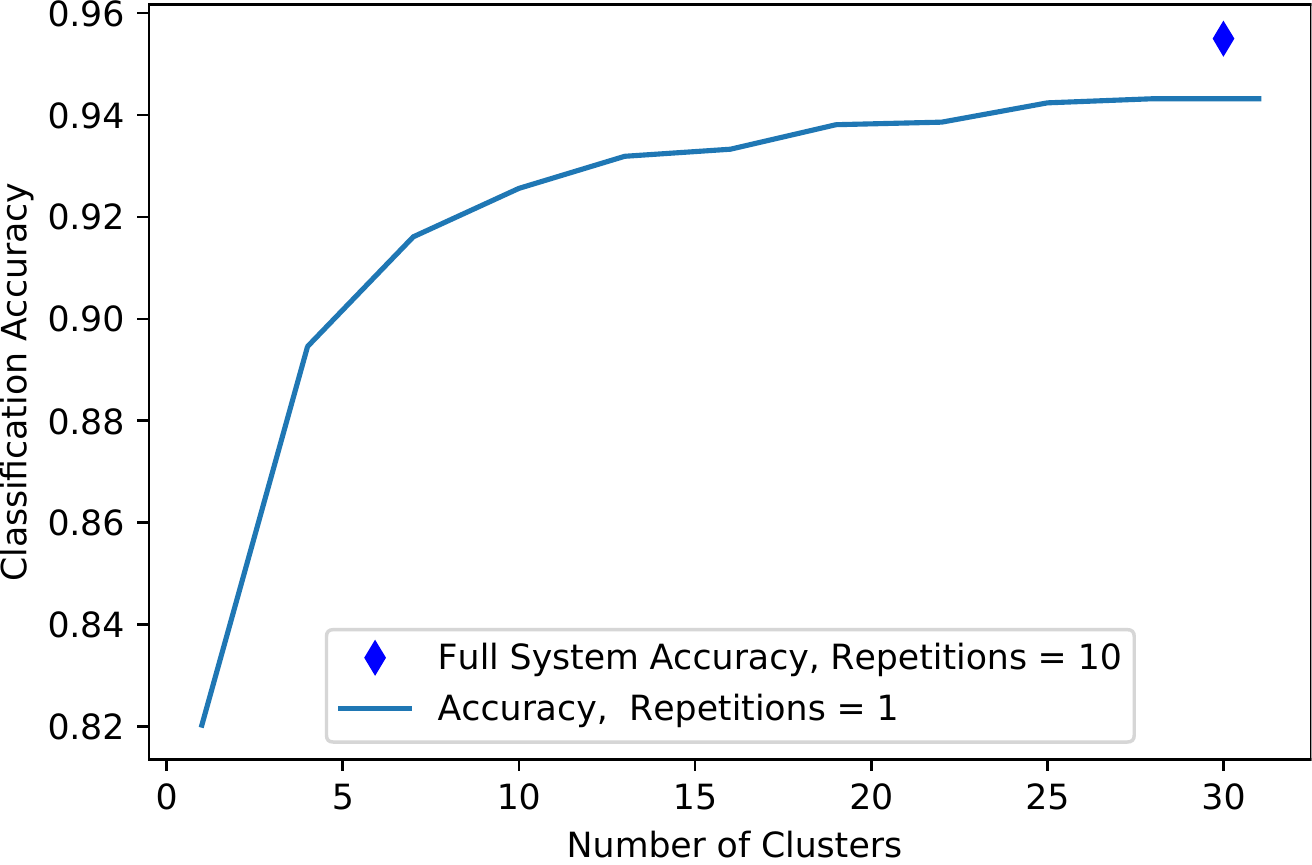} &
 \includegraphics[width=0.4\linewidth]{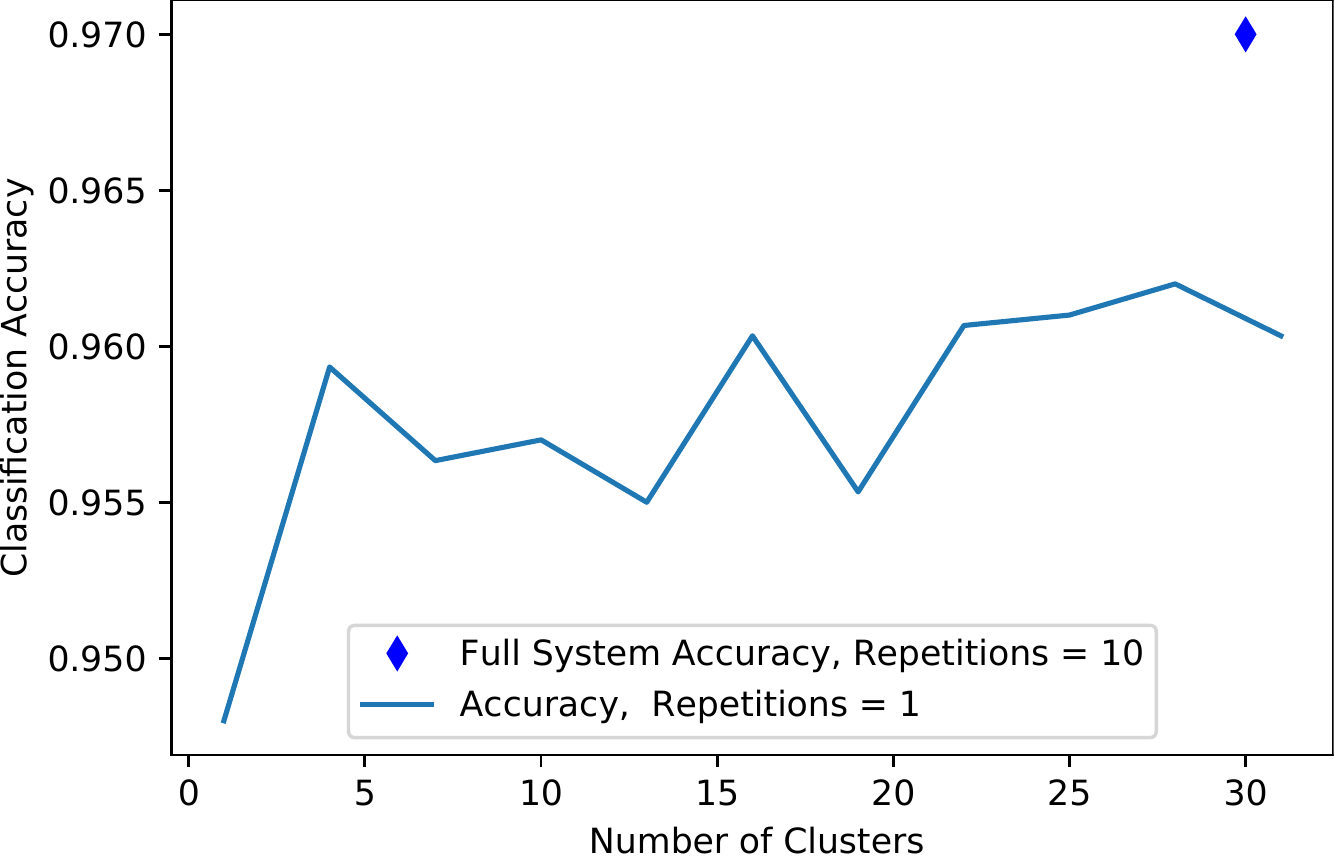}

\end{tabular}
\caption{Using more clusters allows for higher accuracy at the expense of larger representation sizes. Large
representation sizes are less of a problem for distance classifiers because, if necessary,  they can take advantage of 
approximate nearest-neighbor algorithms, as shown in~\fref{fig:time}. 
  \label{fig:ablation}}
\end{figure*}

\Fref{fig:ablation} shows 
the accuracy benefits that accrue to using a large number 
of clusters as well as how adding repetitions gives an additional performance
boost. To maximize accuracy, 
 we represent each class with  $300$ cluster-means (based on $30$ clusters and $10$ repetitions). 
 
 Such a large representation sizes are a-typical because 
 traditional classifier perform inference with a soft-max layer,
which has an inference complexity of $O(N)$, where $N$ is the representation size.
In contrast, a distance  classifier can use approximate 
nearest neighbor discovery,  which has  an 
   inference complexity of $O(log(N))$, making representation sizes  less of a concern
   as shown in~\fref{fig:time}.

In practice, we  find that inference speed is  not
a bottleneck in most current datasets, as shown in \tref{tab:testTime};
and thus, not a current concern. However, the lower
complexity of distance classification opens up the possibility of 
to very large inference engines  encompassing millions of classes.

\section{Distance Top-N Classifier (Inference)} 
\label{sec:top-N}

Top-N classification is the   task of
finding the  $N$ best labels
for a  given feature, $\bx(t)$.

Let $\mathbb{M}_i$ denote the set of generator-means used to ``describe'' label-$i$.
 $\mathbb{M}_i$ can be estimated using Algorithm \ref{algo:clustering} in \sref{sec:mini}.
$\mathfrak{M}$ is the set of all estimated generator  means and
 $\mathbb{L}$ is the  set of possible labels.
\begin{equation*}
\mathbb{M}_i \subset \mathfrak{M}; \quad \quad \quad \mathbb{M}_i \cap \mathbb{M}_j = \varnothing, \quad  \forall\,  i \neq j \in \mathbb{L}.
\end{equation*}

The squared distance of $\bx(t)$ from label-$i$ is  
 the smallest squared distance from 
$\bx(t)$ to  any generator-mean associated with label-$i$:
\begin{equation}
\label{eq:dist_to_label}
min_i(t) =   \min_{\bmu \in \mathbb{M}_i} \{ \|(\bx(t) - \bmu\|^2\}.
\end{equation} 

From  \eref{eq:noise_ncmc}, we know that   $\bx(t)$ is most closely related to the nearest generator mean (and thus its
semantic label). Therefore,  $\bx(t)$'s top label is:
\begin{equation}
y_1(t) = \argmin_{i \in \mathbb{L}} min_i(t).
\end{equation}
The top $N^{th}$ label,  $y_N(t)$, is estimated with the same algorithm, modified to exclude the previous top $N$-$1$ labels from consideration. 

\begin{algorithm}[t!]
	\caption{{\bf Learning   Label-$\mathbf{i}$} \label{algo:clustering}}
	\small
	\textbf{Input:} \\ 
	$K = 30\,;$ \quad \# number of clusters  \\ 
	$R = 10\,;$ \quad \# number of repetitions\\
	$\mathcal{X}_i\,.$ \quad \# set of  features from class $i$ \\

	\texttt{\\}
	\textbf{Initialization:} \\ 
    $ r = 0\,;$\\
	$\overline{\bx}_i$ = mean($\mathcal{X}_i$)\,; \\
    $\widehat{\mathcal{X}_i} = \left\{ \left. \hat{\bx}(t) =  \frac{\bx(t)-\overline{\bx}_i}{\|\bx(t)-\overline{\bx}_i\|}  \,\right|\, \bx(t) \in \mathcal{X}_i \right\}; $\\	
    $\mathbb{M}_i = \{\}\,.$
	
	\texttt{\\}
	\textbf{Estimate generator-means:} \\ 
	\While { $r < R$ :}{
	 \# $\mathcal{C}_k$ is the set of  indices belonging to cluster $k$ \#\\
		$\{\mathcal{C}_k\}$ = K-means-cluster($\widehat{\mathcal{X}_i}, K$)\\
		\For { $k < K$ :}{
			$\bm_k$ = mean($\{\bx(t) \,|\,  t \in \mathcal{C}_k\}$)\\
			$\mathbb{M}_i = \mathbb{M}_i  \cup \, \{\bm_k\} $\\

    		}
	$r = r +1$
    }
    \textbf{return} $\mathbb{M}_i\,.$
	
	\texttt{\\}   
    \# an optional  step for label centric validation; \#\\
    \# not used in top-N classification.\#\\
    	\textbf{Additional Input:} 
	$\,\bm\,.$ \quad \#  common mean of  data\\
    From  \eref{eq:per},  estimate histogram, $h_i(d, \bm)$\\
    \textbf{return}  $h_i(d, \bm)\,.$

\end{algorithm}

\begin{table*}[htp!]
	\footnotesize
	\centering
	\begin{tabular}{l|l|c|c|c|c|c|c}
		\toprule
		&&\multicolumn{6}{c}{\textit{Average classification accuracy on each  dataset}}\\
		\hline
		&Algorithm & MNIST &\makecell{\small Fashion-MNIST}  &\makecell{STL-10}& \makecell{Internet STL-10} & \makecell{SUN-Small} & ASSIRA \\
		\hline
	    \multirow{3}{*}{\makecell[l]{ Discriminatively \\ Trained Classifiers}}    	
		& SVM-linear~\cite{Hearst} & 0.912& 0.810 & 0.944 & 0.880 & 0.926 & \textbf{0.988}\\
		& SVM-kernel~\cite{platt1999probabilistic} & \textbf{0.966} & \textbf{0.884} & \textbf{0.969} & \textbf{0.915} & \textbf{0.939} & 0.987\\
		& Soft-Max Layer & 0.897 & 0.810 & 0.967 & 0.897 & 0.926 & \textbf{0.988}\\

		\hline
		\multirow{ 3}{*}{\makecell[l]{Generative \\ Naive Bayes Classifiers}}    
   		&	GaussNB~\cite{zhang2005exploring}    & 0.556 & 0.586  & 0.926          & 0.864          & 0.877      & 0.981\\
 		&	MulitNomialNB~\cite{schutze2008introduction} & \textbf{0.837} & \textbf{0.655} & \textbf{0.941} & \textbf{0.895} & \textbf{0.913} & \textbf{0.983}\\
		&	ComplementNB~\cite{rennie2003tackling}  & 0.729 & 0.606  & 0.936          & 0.846          & 0.895      & 0.983\\ 

		\hline
		
		\multirow{2}{*}{\makecell[l]{Weakly Incremental \\ Learners}}  
		& OCSVM-N~\cite{chen2001one,lin21} & 0.739 & 0.617  & 0.924 & 0.862 & 0.865 & 0.984\\
		& Shell-N~\cite{lin21} & \textbf{0.824} & \textbf{0.681} & \textbf{0.945} & \textbf{0.901} & \textbf{0.903} & \textbf{0.985}\\

\hdashline  	    

		 \multirow{4}{*}{\makecell[l]{Strictly Incremental  \\ Learners}} 	
  	    & OCSVM-R~\cite{chen2001one} & 0.100  &  0.101 & 0.735 & 0.820 & 0.656 & 0.958\\ 
  	    	& Shell-R & 0.820  &  0.676 & 0.930 & 0.902 & 0.911 & 0.981\\
  	    	& Mahalanobis~\cite{lee2018simple} & 0.753 & 0.677 & 0.922 & 0.453 & 0.770 & 0.979\\

	& Distance (top-N)  & \textbf{0.955} & \textbf{0.825} & \textbf{0.970} & \textbf{0.907} & \textbf{0.929} & \textbf{0.986}\\

		\hline
	\end{tabular}
	
	\caption{Classification accuracy on traditional benchmarks. 
	Distance  (top-N) is a strict incremental learner  with an accuracy that is more commonly associated with  discriminative classifiers. 
	 \label{tab:mc}}

\end{table*}

\subsection{Small Datasets}
\label{sec:small_mult}

We begin the evaluation on  benchmarks familiar to many researchers. They are:
\begin{itemize}
\item MNIST~\cite{lecun1998gradient}\,:  the classic  dataset of handwritten numbers ranging from zero to nine.
\item Fashion-MNIST~\cite{xiao2017fashion}\,: consists of an assortment of clothing silhouettes grouped into $10$ different labels.
\item STL-10~\cite{coates2011analysis}\,:  contains images from $10$  commonly occurring  labels. For example,  bird, cat and  truck.
\item Internet STL-10~\cite{lin21}\,:  shares the same  labels as STL-10; however, training data is crawled from the internet and contains many
outliers; testing is conducted on the original STL-10.
\item SUN-Small~\cite{zhou2014learning}\,:  comprises of the first five labels of SUN-397~\cite{zhou2014learning}. These are: abbey, alley, airport terminal, amusement park and aquarium. 
\item ASSIRA~\cite{asirra}\,:  contains two, highly ambiguous labels, dog and cat.  
\end{itemize}

 MNIST and Fashion-MNIST images are represented by pixel  rasterization. This
is possible because their  objects of interest  are simple, appropriately scaled and centered in each image.
 For the other datasets, images  are represented by  ResNet-50~\cite{he2016deep} features,
  whose  weights were pre-trained on  imageNet~\cite{deng2009imagenet}.
  The deep-learned feature representation is necessary because  objects of interest are complex and unaligned, making
  pixel based image comparison   meaningless.

Classification algorithms  are trained on a subset of each dataset and evaluated on an unseen  test set.
Classification accuracy is reported in  \tref{tab:mc}. Algorithms 
are divided into three categories: discriminative classifiers, generative classifiers
and incremental learners. The highest accuracy for each category is  bolded.

  Discriminative classifiers are 
represented by   linear-SVM~\cite{Hearst}, kernel-SVM~\cite{platt1999probabilistic} and 
soft-max classifiers.   Unsurprisingly, 
 discriminative classifiers are the best performing category of classifiers.

Generative classifiers are represented by variants of the popular  Naive  Bayes framework~\cite{zhang2005exploring,schutze2008introduction,rennie2003tackling}. They
 employ joint training to learn the identifying statistics of each semantic. 
As expected, the generative  classifiers are significantly less accurate  than  
discriminative classifiers.  

The final set of algorithms are incremental learners which  
 learn each semantic independently.
 Incremental learning algorithms  are further  divided into two groups.  The first 
comprises of  algorithms that use  shell  normalization~\cite{lin21} (see  \sref{sec:norm} for details).
Shell normalization  significantly enhances accuracy~\cite{lin21}; however, it requires the dataset mean, which 
is   information external to a semantic. We regard algorithms that use shell normalization, to have 
weakly fulfilled the   incremental learning requirements. 
The second group of incremental learners employ no normalization  and are deemed  strict incremental learners. As expected, incremental learners
have generally low  accuracy, with  strict incremental learners being the most inaccurate.\footnote{ Our  evaluation  focuses only on the top one-class learning
 algorithms for these datasets. Comparisons with other one-class learning algorithms can be found in~\cite{lin21}. 
}

 Distance (top-N)  inverts  the  performance hierarchy.
 Although Distance (top-N)  is  a strict incremental learner,  its 
  accuracy approaches the upper ranks of   discriminative classifiers.
  This is an exciting prospect because  Distance (top-N)'s provides significant 
  advantages in training complexity and ease of update.

\subsection{Training Complexity}
\label{sec:speed}


Traditional machine learning uses the training step to discover     classification constraints.
In distance based  classification, the  constraints  are analytically derived prior to training; this significantly reduces
  training complexity.

Let $N$ denote the number of semantic labels, the training complexity of  Distance (Top-N) is $O(N)$ or $O(Nlog(N))$. 
The $O(N)$ complexity comes about because  Distance Classifiers  learn
each semantic independently. A complexity of   $O(Nlog(N))$  is incurred if an
additional search  tree  is constructed~\cite{Hyvonen2016,Jaasaari2019}.
Distance (Top-N) has a test complexity of   $O(log(N))$ if a search tree is used and  $O(N)$ otherwise. 
Distance Classification's  very low training and test complexity makes it scalable to large datasets. 
In contrast, the   training and  test complexity of linear-SVM, one of the fastest traditional classifiers, is $O(N^2)$ and   $O(N)$    respectively.

\begin{figure*}[htp!]
\centering
\begin{tabular}{cc}
Training Complexity & Test Complexity\\
\includegraphics[width=0.48\linewidth]{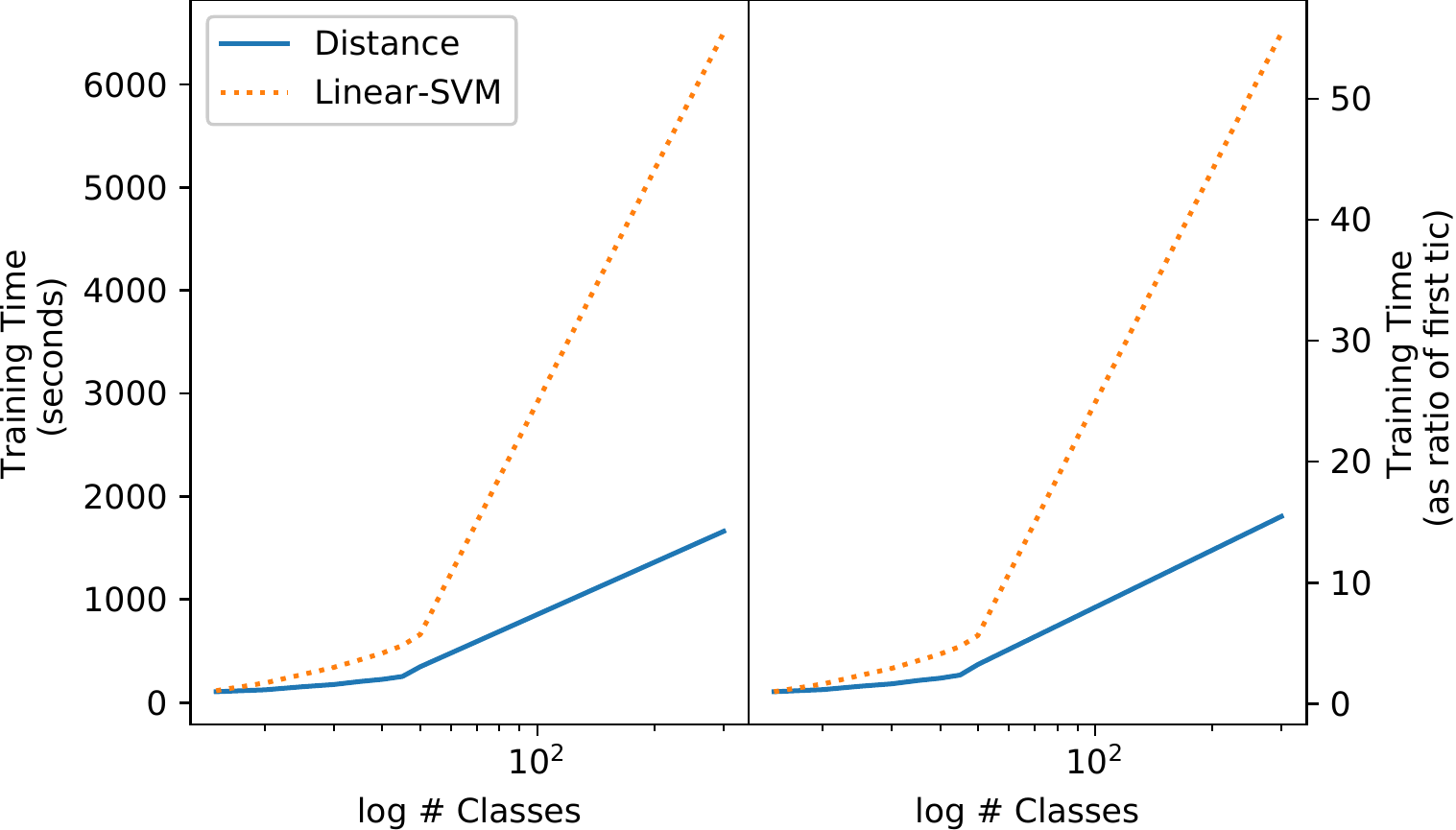}& 
\includegraphics[width=0.48\linewidth]{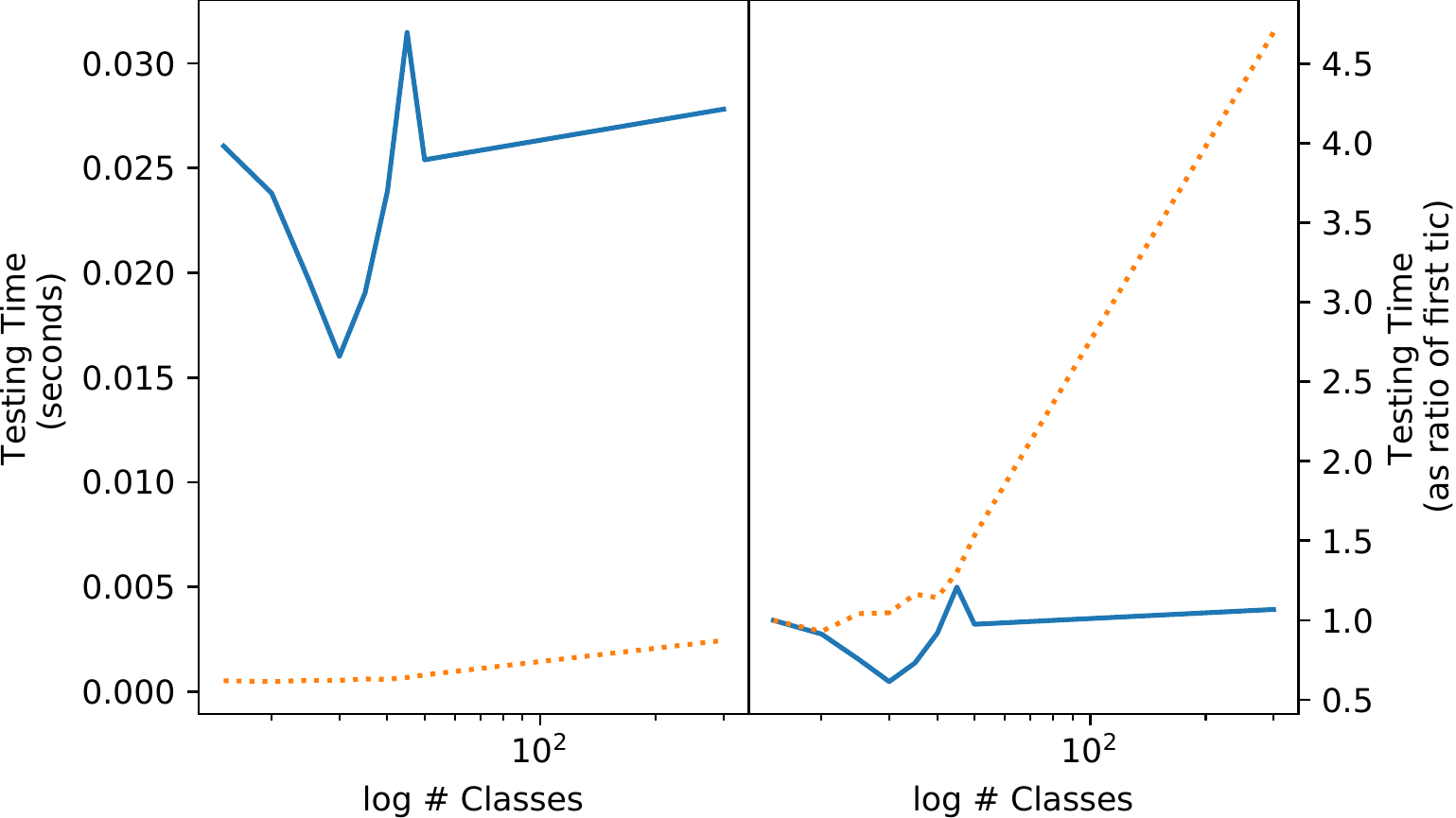}
\end{tabular}
\caption{Comparing training and testing speeds with increasing  number of classes. Computational time 
is measured in seconds and as a ratio of the computational time at the first tic mark on the graph (15 classes). 
Distance Classifiers have a lower    \textbf{training complexity} than  linear-SVM,
$O(Nlog(N))$ vs $O(N^2)$, where $N$ is the number of labels.   Distance Classifiers
also have a lower
 \textbf{test complexity}, $O(log(N))$ vs $O(N)$. Although 
  the Distance Classifier's  initial test time is higher than linear-SVM; the 
  test time grows very slowly with increasing number of classes.  
 \label{fig:time}
}
\end{figure*}

\Fref{fig:time} shows  real-world training and test times on SUN-397~\cite{zhou2014learning}. 
Distance (Top-N)'s training is clearly faster than  linear-SVM's,  with the gap  growing
rapidly as the number of classes increase.  Distance (Top-N)'s absolute test time is slower than linear-SVM's;  however, Distance (Top-N)'s test time  grows more slowly.

\subsection{Incrementally Learned  Large Datasets}

The strictly incremental, one-class learners in \sref{sec:small_mult}
are almost never employed on large scale datasets because their
performance drops rapidly as the number of classes increase. Indeed,
the consensus seems to be that 
strictly incremental learning is impractical
and most 
 large scale 
incremental learners use some      replay of past data  in their update steps~\cite{rebuffi2017icarl,castro2018end,wu2019large,hayes2019remind,rajasegaran2020itaml}.
In this paper's definition, such  algorithms are  weakly incremental.

To ensure efficient updates,   weakly incremental  try to
  minimize replay. However, even if  
one instance per semantic is selected for replay,  the update complexity 
of a weakly incremental learner is  at least
$O(N)$, where $N$ is the number of previously learned   semantics.
To manage computational cost,   
these  learners employ simpler neural-networks, like ResNet-18~\cite{he2016deep},
trading accuracy for speed. 

Distance  (top-N) provides  a strictly incremental learner which does not require replay.
As each  semantic is learned independently, the update complexity of Distance  (top-N) is  $O(1)$, or $O(log (N))$ if a  tree is created  for fast search.  The increased update speed
makes it practical to use   
better features, such as ResNet-50~\cite{he2016deep} (instead of ResNet-18), which in turn improves accuracy. 

The disadvantage of  Distance  (top-N)  is it lacks a mechanism  for   feature refinement.
Instead,   Distance  (top-N)  relies on  noise cancellation  to accommodate   less than ideal features.
Nonetheless, the overall system is
faster (as shown in \sref{sec:inc_speed})
 and more accurate (as shown in \tref{tab:evalAll}) than traditional incremental learners.   Below, we
 discuss the results in   detail.
\newline

\noindent\textit{Baseline algorithms~\cite{wu2019large,hayes2019remind,rajasegaran2020itaml}:} These 
are  recent, top performing, large scale incremental learners.
Baseline algorithms enjoy the benefit of  replay, which is not used in   distance classification.
  To further ensure a fair comparison, baseline algorithms are  initialized 
with pre-trained imageNet weights, a setting which we find maximizes their  performance.   
\newline

\noindent\textit{ImageNet-1000~\cite{deng2009imagenet}:}  This is the traditional benchmark for classification algorithms. Distance
classification's  top-5 accuracy on imageNet is $88\%$, which is quite close   to  the   $94.7\%$ accuracy of  jointly-trained  ResNet-50~\cite{he2016deep},  the  upper bound for an 
  incremental learner's performance~\cite{wu2019large}.  
 ImageNet's test set has   many erroneous labels, some of which have  been  corrected~\cite{beyer2020we}.  
Evaluated on the new test labels, distance classification's  top-5  accuracy  improves  to $91.7\%$, while  the accuracy of ResNet-50 declines slightly~\cite{beyer2020we}, further narrowing  the performance gap.   

As the  ResNet-50 features used in distance classification
were learned using  imageNet,  distance classification's performance  on imageNet 
should  not be considered a measure  of it's  generalizability. 
Instead, the imageNet evaluation should be interpreted as a test for    distance classifier's
ability to   replace  ResNet-50's final soft-max layer.
From this perspective, the results on imageNet are  promising. 
To evaluate general classification accuracy, we turn to the next three datasets. 
\newline

\noindent \textit{SUN-397~\cite{xiao2010sun}, Food-101~\cite{bossard2014food}, Flower-102~\cite{nilsback2006visual}:} These datasets
consist of buildings, food and flowers respectively.
As these  semantic concepts    are very different from  the animal centric semantics of imageNet,
the classes can be considered ``unseen'' by  ResNet. These datasets are very challenging 
and  even  state-of-the-art incremental learners exhibit significant performance 
variations, with the  top performer on one dataset,
sometimes failing  completely on a different dataset. In contrast,
Distance (Top-N)  is  consistently  the best algorithm or a  close second.

\begin{table*}[!htp]
	\begin{minipage}{1\linewidth}
		\renewcommand{\tabcolsep}{2.5pt}
		\renewcommand{\arraystretch}{1.2}
		\centering
		\begin{tabular}{l|l|ccccccccccccc}
			\multicolumn{13}{c}{\textbf{\textit{Top-5 accuracy (\%)}}}\\
			\toprule
			\multicolumn{2}{r}{Increment:} & 1 & 2 & 3 & 4 & 5 & 6 & 7 & 8 & 9 & Final  &  Decline \\
			\hline
			\multirow{3}{*}{\textbf{ImageNet-1000}} &BiC~\cite{wu2019large} & 94.1 & 92.5 & 89.6 &89.1 &85.7 &83.2 &80.2 &77.5 &75.0 &73.2 & 20.9\\ 
			\multirow{4}{*}{{100 class / inc}}&REMIND~\cite{hayes2019remind} & 94.7 & 87.1 & 83.5 & 80.1 & 77.1& 75.3&74.1 &72.9 & 72.4 &71.2 & 23.5\\
			&iTAML~\cite{rajasegaran2020itaml}  & 91.5 & 89.0 & 85.7 & 84.0 & 80.1 & 76.7 & 70.2 & 71.0 & 67.9 & 63.2 & 28.3\\

			&Distance           & 94.0 & 94.0 & 93.2 & 92.7 & 92.8 & 92.1 & 90.8 & 89.5 & 88.2 & \textbf{88.0} & 6.0\\
			&\makecell[l]{Distance* \\(relabeled)}  & 95.6 & 95.6 & 95.1 & 94.7 & 94.1 & 93.2 & 92.4 & 91.8 & 91.3 & 91.7 & 3.9\\

			\hline
			\multirow{3}{*}{\textbf{SUN-397}}    &BiC~\cite{wu2019large} & 84.1 & 76.7 & 61.7 &  50.4 & 48.3 & 47.2 &34.2 & 30.3 & 24.3 &22.6  &  61.5\\
			\multirow{3}{*}{40 class / inc}&REMIND~\cite{hayes2019remind} & 92.9 & 88.7 & 85.9 & 83.0 & 81.2 & 77.8 &77.9 &74.8 &73.8 &72.6 &20.3\\
			&iTAML~\cite{rajasegaran2020itaml}  & 95.8 & 90.1 & 86.9 & 81.9 & 78.6 & 75.7 & 75.3& 73.3& 71.9&68.9 &  26.9\\

			&Distance  & 99.4 & 97.8 & 95.8 & 94.4 & 92.9 & 92.0 & 90.2 & 88.8 & 88.1 & \textbf{87.5} & 11.9\\
			\hline
			\multirow{3}{*}{\textbf{Food-101}}    &BiC~\cite{wu2019large} & 88.2 & 82.0 & 80.3 & 77.9 & 72.4  & 73.5 & 69.1 & 68.3. & 65.6 &61.5  &26.7 \\
			\multirow{3}{*}{10 class / inc}&REMIND~\cite{hayes2019remind} & 92.3 & 91.1 & 87.6 & 85.2 & 83.6 & 81.0 &81.1 &79.4 &78.2 &75.9 &16.4\\
			&iTAML~\cite{rajasegaran2020itaml}  & 98.6 & 97.0 & 95.2 & 94.3 & 92.8 & 91.1 & 89.4& 87.5& 86.6&\textbf{84.7} &  13.9\\
			& Distance   & 99.0 & 93.0 & 91.0 & 88.8 & 87.7 & 85.8 & 85.3 & 84.5 & 82.9 & 81.1 &  17.9\\
			\hline
			&    & 1 & 3  & 5 &  7 &  9 &  11 &  13 &  15 &  Final && Decline\\
			\hline
			\multirow{3}{*}{\textbf{Flower-102}}    &BiC~\cite{wu2019large} 
			&92.1 &65.4 &53.1  &39.6  &32.0  &38.7  &31.0  &30.1  &35.9  && 56.2\\
			\multirow{3}{*}{6 class / inc}&REMIND~\cite{hayes2019remind} 
			&100 &  99.0&  99.4&  98.8&  98.7& 99.1&  98.9&  97.4&  96.8 & &3.2\\
			&iTAML~\cite{rajasegaran2020itaml}  
			& 100 &  94.4  & 92.8  & 88.8&  90.0&  88.4  &85.6  &83.2 & 81.0 & &19.0 \\
			& Distance  &100  & 100  & 100  & 99.5  & 98.7  & 99.0  & 98.1  & 97.3  & \textbf{97.1} && 2.9\\
			\hline
		\end{tabular}
		
		\vskip 0.1cm
		\begin{tabular}{l|l|ccccccccccccc}
			\multicolumn{13}{c}{\textbf{\textit{Top-1 accuracy (\%)}}}\\
			\toprule
			\multicolumn{2}{r}{Increment:} & 1 & 2 & 3 & 4 & 5 & 6 & 7 & 8 & 9 & Final  & Decline\\
			\hline
			\multirow{2}{*}{\textbf{ImageNet-1000}}  

			& Distance 
			& 76.7 & 76.7 & 74.3 & 73.9 & 74.9 & 73.8 & 71.9 & 70.3 & 68.3 & 67.7 & 9.0\\
			\multirow{1}{*}{100 class / inc}
			& \makecell[l]{Distance*\\ (relabeled)}  & 81.7 & 81.2 & 79.0 & 78.8 & 78.4 & 77.0 & 75.8 & 74.8 & 73.9 & 74.2 & 7.5\\
			\hline
			
			\multirow{2}{*}{\textbf{SUN-397}}  &REMIND~\cite{hayes2019remind} & 71.8& 66.9& 61.6 & 57.0& 54.6& 49.6& 51.8& 47.7& 47.7& 46.7 & 25.1\\
			\multirow{2}{*}{40 class / inc}
			&iTAML~\cite{rajasegaran2020itaml}  & 65.7 & 53.1 & 44.1 & 36.3 & 34.1 & 32.0 & 28.9& 25.4& 23.5& 21.6&  44.1\\
			&Distance  & 86.6 & 81.1 & 74.7 & 71.4 & 68.3 & 66.4 & 64.0 & 61.8 & 61.2 & \textbf{60.6} & 26.0 \\
			\hline
			\multirow{2}{*}{\textbf{Food-101}} &REMIND~\cite{hayes2019remind} &66.3&64.1&61.9 & 58.9& 57.6 & 58.1 & 55.4 &54.1&50.9&50.6 &15.7\\
			\multirow{2}{*}{10 class / inc}
			&iTAML~\cite{rajasegaran2020itaml}  &87.5 & 82.1 & 77.8& 74.5& 72.5& 67.8 & 62.6& 59.6& 58.6& \textbf{57.9}& 29.6 \\
			&Distance   & 74.9 & 65.3 & 65.0 & 63.7 & 61.8 & 59.8 & 60.0 & 59.3 & 57.0 & {55.2} & 19.7\\
			\hline
			&    & 1  & 3  & 5  & 7 & 9 &  11 &  13 &  15  & Final && Decline\\
			\hline
			\multirow{2}{*}{\textbf{Flower-102}} 
			&REMIND~\cite{hayes2019remind} &100 &  92.1&  93.6&  91.7&  88.6&  92.3& 90.8&  87.7&  85.3 && 14.7\\
			\multirow{2}{*}{6 class / inc}
			&iTAML~\cite{rajasegaran2020itaml}  
			&88.1  &66.4  &66.5  &61.5  &57.9  &53.7  &48.2 &43.4&41.0  && 47.1\\
			&Distance  &100&  96.2 & 96.5  & 94.6  & 92.8  & 92.5 &  90.7 &  88.6  & \textbf{87.6} &&12.4 \\
			\hline
		\end{tabular}
		\caption{ Top-5 and Top-1 accuracy (\%) of incremental learners.  Decline is the accuracy difference   between the first and  last increment.  ImageNet results are duplicated  from the  original papers. Distance classification's Top-1 accuracy is $11.3\%$ better than REMIND, its closest competitor.  \label{tab:evalAll}}
	\end{minipage}\hfill 
\end{table*}

\subsection{Speed Comparisons}
\label{sec:inc_speed}

 Distance (Top-N) can also be trained efficiently,  
as shown in \tref{tab:update}. The performance gain being especially noticeable on large 
datasets like SUN-397.  Further, although Distance (Top-N)  uses
a more complex representation than the soft-max layer of other incremental leaning,
its inference time remains low, as shown in \tref{tab:testTime}. If additional inference speed  is required,
the  approximate nearest neighbor algorithm, discussed in  \sref{sec:speed}, can be employed.

	\begin{table*}[h!]
	\center
	\scriptsize
	\begin{tabular}{l|l|c|c|c|c}
	\hline
	Dataset & Method & \makecell{Num. labels learned\\ before last increment} & \makecell{Num. labels to learn \\ in last increment} 
	& \makecell{Training time for\\  last increment (seconds)} & \makecell{Training time \\  per label (seconds)} \\
	\hline
	\multirow{4}{*}{\textbf{SUN-397}}    
	& BiC~\cite{wu2019large} 						& 360 & 37 & 66002 & 1784\\
	& REMIND~\cite{hayes2019remind}               	& 360 & 37 & 2452  & 66\\
	& iTAML~\cite{rajasegaran2020itaml}           	& 360 & 37 & 73601 & 1989 \\
	& Distance  								    & 360 & 37 & \textbf{453}   & \textbf{12.2}\\
	\hline
	\multirow{4}{*}{\textbf{Food-101}}    
	& BiC~\cite{wu2019large} 						& 90 & 11 & 1524 & 139\\
	& REMIND~\cite{hayes2019remind}               	& 90 & 11 & 1090 & 99\\
	& iTAML~\cite{rajasegaran2020itaml}           	& 90 & 11 & 1553 & 141 \\
	& Distance  									& 90 & 11 & \textbf{498}  & \textbf{45.3}\\
	\hline
	\multirow{4}{*}{\textbf{Flower-102}}    
	& BiC~\cite{wu2019large} 						& 90 & 12 & 627 & 52\\
	& REMIND~\cite{hayes2019remind}               	& 90 & 12 & \textbf{19} & \textbf{1.6}\\
	& iTAML~\cite{rajasegaran2020itaml}           	& 90 & 12 & 1248 & 104 \\
	& Distance  									& 90 & 12 & 38 & 3.5 \\
    \hline

	\end{tabular}
	\caption{Time to incremental update a  representation. Distance classification has a complexity advantage over 
	other algorithms and  thus, is notably faster  on large datasets like SUN-397.   \label{tab:update}}

		\begin{tabular}{l|c|c|c}
	\hline
	 & SUN-397 & Food-101 & Flower-102 \\
	\hline
	\makecell{Distance \\ Time to classify an image\\(seconds)} & 0.014 & 0.007 & 0.002\\
    \hline

	\end{tabular}
    \caption{Time taken for distance classifier   to classify an image through exhaustive testing of all means; the GPU used is a RTX 2080Ti. As classification times are consistently   below $0.1$  seconds,  testing time is not a bottleneck.  \label{tab:testTime} } 
	\end{table*}

\section{Label Centric   Validation (Inference)} 
\label{sec:val}

The top-N classification  discussed in  \sref{sec:top-N}  suffices
if each feature has one and only one label.
However,  labels often overlap one another and features may derive from 
labels  outside the  training set. Thus, rather than  finding  the best label for a given feature, label centric validation
 seeks to  estimate the likelihood of  a   feature
belonging  to a  given label.

\subsection{Distance Label Validation}

As explained in \sref{sec:noise_ncmc} and \fref{fig:strange},
in the presence of noise, the raw distance of a feature to a set of means, is a good classification metric
but a terrible validation metric. 
We propose two metrics for label centric validation;
the first is based on distance ratios, which are shown in \sref{sec:ratio_cancel}  to be noise invariant;
the second is based on noise cancelled distance of features to means, derived in \sref{sec:ref_cancel}. 
\newline

\noindent\textit{Ratio Based  Validation:}
Ratio based validation is based on the intuition that 
a features should be unusually close    to the label-$i$ before
it is  validated by label-$i$.

We use the same  notation as in  \sref{sec:top-N}, 
$\mathfrak{M}$ is the set of all estimated generator-means (\sref{sec:mini}, Algorithm \ref{algo:clustering})
and  $\mathbb{M}_i \subset \mathfrak{M}$  is the set of generator  means associated with label-$i$. 

For  convenience, we  define the following entities:
\begin{equation}
\begin{split}
 & min_i(t) &=   \min_{\bmu \in \mathbb{M}_i} \{ \|(\bx(t) - \bmu\|^2\},\\
 &min\_(t) &=  \min_{\bmu \in \mathfrak{M}} \{ \|(\bx(t) - \bmu\|^2\},\\
 &max\_(t) &=  \max_{\bmu \in \mathfrak{M}} \{ \|(\bx(t) - \bmu\|^2\}.\\
 \end{split}
 \end{equation}
$min_i(t)$ is $\bx(t)$'s squared distance to the closest generator-mean in label-$i$, a definition   repeated from \eref{eq:dist_to_label}.
$min\_(t)$ is $\bx(t)$'s squared distance to the closest generator-mean in $\mathfrak{M}$.
$max\_(t)$ is $\bx(t)$'s squared distance to the further generator-mean in $\mathfrak{M}$.

The small-ratio scores   $\bx(t)$ by how close it is  to label-$i$,  relative to  $\bx(t)$'s  closest mean:
\begin{equation}
\label{eq:small}
s_i(t) = \frac{min_i(t)} {min\_(t)};
\end{equation}
the smallest (and best) possible  value for $s_i(t)$ is  $1$. 

The large-ratio scores  $\bx(t)$   by how close it is to label-$i$ relative to  $\bx(t)$'s  furthest mean:
\begin{equation}
\label{eq:large}
l_i(t) = 1 + \frac{min_i(t)} {max\_(t)};
\end{equation}
the smallest (and best) possible  value for $l_i(t)$ is $1$.

\Sref{sec:ratio_cancel} shows the noise invariance of  ratios is  limited to similar distances.
To have a metric which spans the  range of possible distances,  
the final ratio metric is a composite of both  small and large ratios: 
\begin{equation}
\label{eq:sl_ratio}
r_i(t)= s_i(t)\times l_i(t).
\end{equation}
\newline

\noindent\textit{Distance  Based  Validation:} 
\Eref{eq:noise_cancel1} shows that the noise cancelled distance
of a feature to a generator-mean  is a measure of  how closely they are related. 
Distance based validation algorithmizes  this property. 

Let  $\bm$  denote  the reference vector used in noise cancellation. As explained in   \sref{sec:ref_cancel},
an ideal choice of $\bm$ is  the mean of the dataset.
If this cannot be determined, $\bm$  can be set to the mean of a random image collection like Flickr11k~\cite{kuo2011unsupervised}.

From \sref{sec:ref_cancel}, the noise cancelled distance from  label-i is denoted:
\begin{equation}
\label{eq:train_noise}
\widetilde{min}_i(t, \bm) = \min_{\bmu \in \mathbb{M}_i}  d^2(\bx(t), \bmu, \bm).
\end{equation}
To convert distance into a likelihood, we use  a
(inverse) percentile  likelihood function:  
\begin{equation}
\label{eq:per}
h_i(d, \bm) = \frac{ \left| \{t  \,|\, t \in \mathcal{T}_i,\, \widetilde{min}_i(t, \bm) >d\}\right|}{|\mathcal{T}_i|}.
\end{equation}
 $|\,.\,|$ denotes set cardinality (number of members  in a set),
$\mathcal{T}_i$ is the set of indices belonging to label-$i$
and $d$ is a squared distance. This makes $h_i(d, \bm)$,  the percentage of instances of label-$i$  whose noise cancelled distance  to label-$i$ exceeds $d$.
The (inverse) likelihood of  $\bx(t)$'s membership with label-$i$ is 
\begin{equation}
\label{eq:likelihood}
a_i(t, \bm) = h_i(\widetilde{min}_i(t, \bm), \bm),
\end{equation}
where $a_i(t, \bm)$ varies between $0$ (best) and $1$ (worst). 
\newline

\noindent\textit{Final Validation Metric:} The final validation metric 
fuses  the ratio  and distance  metrics from  \eref{eq:sl_ratio} and  \eref{eq:likelihood}:
\begin{equation}
\label{eq:final_veri}
v_i(t, \bm) = 1 -  a_i(t, \bm) \times r_i(t),
\end{equation}
where the maximum (and best) score is $1$.  Empirically, we find  
$v_i(t, \bm)>0.9$ is a good threshold for accepting  $\bx(t)$ as a member of label-$i$.  
The validation process is summarized in Algorithm \ref{algo:full}.

Note that \eref{eq:final_veri}  assumes  knowledge of the dataset mean, $\bm$.
This  information  is external to an individual semantic;    thus,
 algorithms based on \eref{eq:final_veri} are only  weak one-class learners. 
 
A strict incremental learner is possible, if only ratio validation is used: 
\begin{equation}
\label{eq:final_veri_}
v_i(t) = 1 -   r_i(t).
\end{equation}
While   slightly inferior to \eref{eq:final_veri}, this is  still  a very good validation function. 

\begin{algorithm}[t!]
	\caption{{\bf Distance Validation } \label{algo:full}}
	\textbf{Training:}\\
	\# $\mathfrak{M}$ is the set of all learned generator-means\# \\
	\# $\mathfrak{H}$ is the set of all learned percentile functions\# \\
	$\mathfrak{M} = \{\}$, \\
	$\mathfrak{H} = \{\}$, \\

	\For { each class, $i$ :}{
		Estimate $\mathbb{M}_i$, $h_i(d, \bm)$ with Algorithm \ref{algo:clustering};\\
		$ \mathfrak{M} = \mathfrak{M} \,\cup\, \mathbb{M}_i $;\\
		$ \mathfrak{H} = \mathfrak{H} \,\cup\, \{h_i(d)\} $;\\
	}
	\textbf{return} $\mathfrak{M}$, $\mathfrak{H}$
 
	\texttt{\\}   
	
	\textbf{Testing:}\\
    \# $t$ is the index of the  test instance\# \\
    \# $\by$ is a one-hot array  representing $t$'s label membership\# \\
    \# $\tau=0.9$ is the acceptance threshold \#\\
    	\For { each class, $i$ :}{
		Estimate $a_i(t, \bm)$, using \eref{eq:likelihood}\;
		Estimate $r_i(t)$, using \eref{eq:sl_ratio}\;
		$v_i(t, \bm) = 1 -  a_i(t, \bm) \times r_i(t).$\\
		\If { $v_i(t) > \tau$ }{
	        $\by[i] = $ True
        }\Else{
    		    $\by[i] = $ False
        }
	}
	\textbf{return} $\by$	
\end{algorithm}

\subsection{Experiments}

\begin{table*}[htp]
	\centering
	\begin{tabular}{l|l|c|c|c|c|c|c}
		\toprule
		&\multicolumn{6}{c}{\textit{Average Area Under Precision-Recall Curve}}\\
		\hline
		&Algorithm & MNIST &\makecell{\small Fashion-MNIST}  &\makecell{STL-10}& \makecell{Internet STL-10} & \makecell{SUN-Small} & ASSIRA \\
\hline
\multirow{ 2}{*}{\makecell[l]{Discriminatively \\Trained Classifiers}}	  
& SVM-linear~\cite{Hearst} & 0.912 &  0.842 & 0.960 & 0.947 & 0.963 & \textbf{0.999} \\
& SVM-kernel~\cite{platt1999probabilistic} & \textbf{0.978} & \textbf{0.903} & 0.980 & 0.933 & {0.967} & \textbf{0.999} \\
& Soft-Max Layer & 0.896 & 0.809 & \textbf{0.995} & \textbf{0.963} & \textbf{0.978} & \textbf{0.999} \\
		\hline
\multirow{ 3}{*}{\makecell[l]{Jointly Learned \\ Naive Bayes\\ Classifiers}}	  
& GaussNB~\cite{zhang2005exploring} & 0.630 & \textbf{0.543} & 0.926 & 0.849 & 0.865 & 0.981\\
& MulitNomialNB~\cite{schutze2008introduction} & \textbf{0.746} & 0.509 & \textbf{0.962} & 0.915 & 0.935 & \textbf{0.987}\\
& ComplementNB~\cite{rennie2003tackling} & 0.679 & 0.503 & 0.985 & \textbf{0.942} & \textbf{0.957} & \textbf{0.987}\\ 
\hline
\multirow{ 3}{*}{\makecell[l]{Weakly Incremental \\ Learners}}	&	OCSVM-N~\cite{chen2001one,lin21} 
& 0.855 & 0.673 & 0.960 & 0.916 & 0.957 & \textbf{0.999} \\
		&Shell-N~\cite{lin21} & 0.845 & 0.662 & 0.952 & \textbf{0.951} & 0.947 & \textbf{0.999}\\
		& Distance (VW) & \textbf{0.960} & \textbf{0.770} & \textbf{0.984} & \textbf{0.951}  & \textbf{0.969} & \textbf{0.999}\\
\hdashline
	\multirow{ 4}{*}{\makecell[l]{Strictly Incremental \\ Learners}}	  
		&OCSVM-R~\cite{chen2001one} & 0.105 & 0.103 & 0.675 & 0.567 & 0.599 & 0.862 \\
		&Shell-R~\cite{lin21} & 0.628  & 0.550 & 0.664 & 0.570 & 0.601 & 0.858\\
		&Mahalanobis~\cite{lee2018simple}  & 0.736 &    0.600       & 0.770 & 0.494 & 0.657 & 0.943\\
		&Distance (VS) & \textbf{0.980} &    \textbf{0.852}       & \textbf{0.984} & \textbf{0.931} & \textbf{0.941} & \textbf{0.985}\\
\hline
	\end{tabular}
	
	\caption{Label centric validation evaluated by AUPRC. Distance Classifiers are comparable to traditional   
    discriminative classifier.  
	 \label{tab:auprc}}

\end{table*}

%
%
%
%
%

\begin{table}[htp]
	\centering
	\begin{tabular}{l|c|c|c|c}
		\toprule
		&\multicolumn{4}{c}{\textit{AUPRC}}\\
		\hline
		Algorithm &\makecell{STL-10}& \makecell{Internet \\ STL-10} & \makecell{SUN\\-Small} & ASSIRA \\
\hline
 NAT~\cite{lu2021neural} & 0.875 & - & - & -  \\
 SVM-linear~\cite{Hearst} & 0.898 & 0.882 & 0.818 & 0.846 \\
 SVM-kernel~\cite{platt1999probabilistic} &  \textbf{0.978} & 0.885 & {0.826} & \textbf{0.991} \\
 Soft-Max Layer& 0.922 & 0.861 & 0.824 & 0.944 \\
 MulitNomialNB~\cite{schutze2008introduction}  & 0.613 & 0.626 & 0.597 & 0.564\\
Distance (VW) &  0.970 & \textbf{0.922} & \textbf{0.852} & 0.979\\

\hline
	\end{tabular}
	
\caption{Comparing classifier's ability to identify novel instances.
Distance classifiers are very  effective; their  AUPRC scores  are consistently   the best or a close second best. \label{tab:novel} }

\end{table}

\begin{figure*}
\centering
\begin{tabular}{cc}

{STL-10} \cite{coates2011analysis} & {SUN-Small}~\cite{xiao2010sun}\\
  \includegraphics[ width=0.48\linewidth]{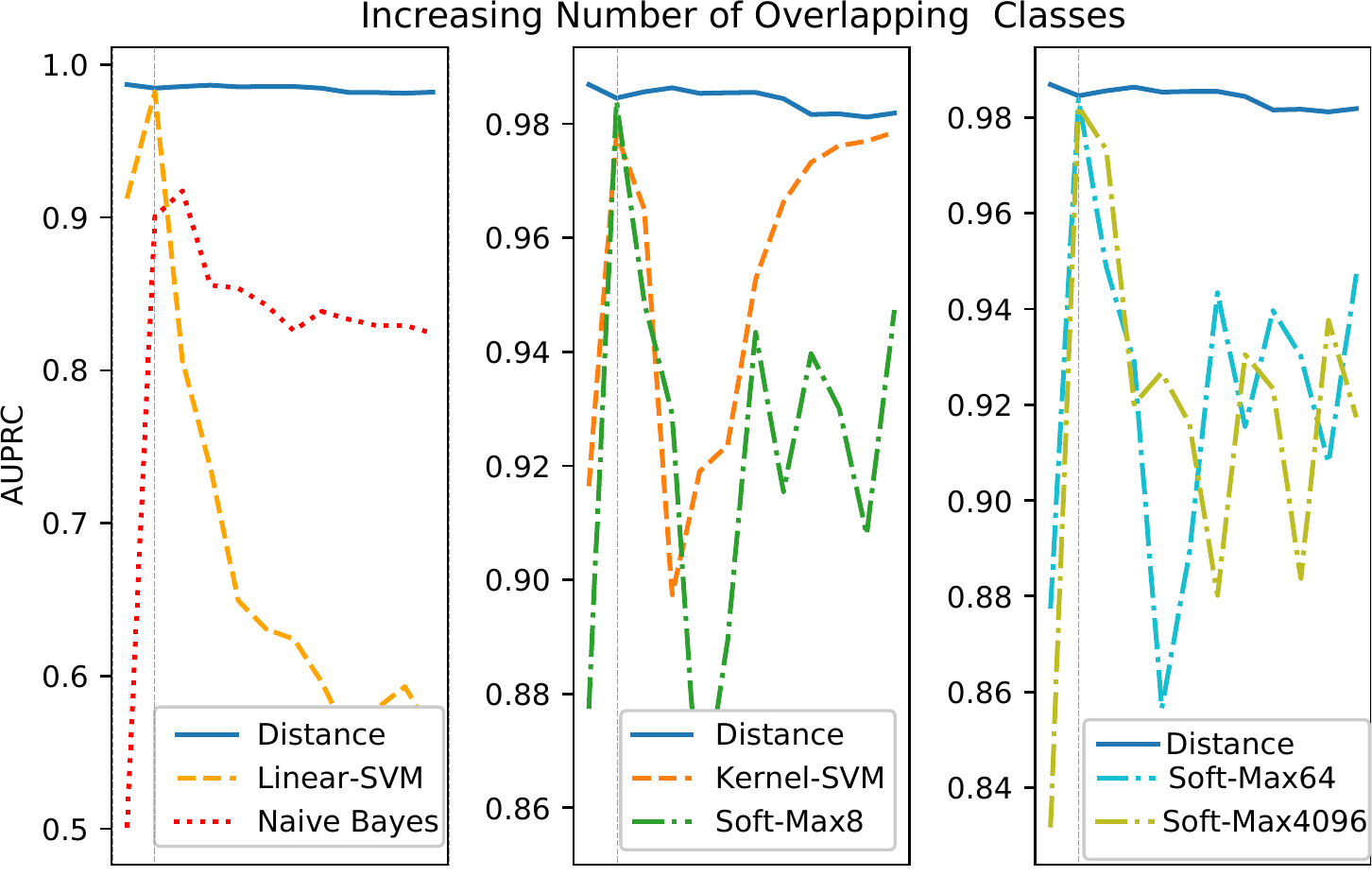}  & \includegraphics[ width=0.48\linewidth]{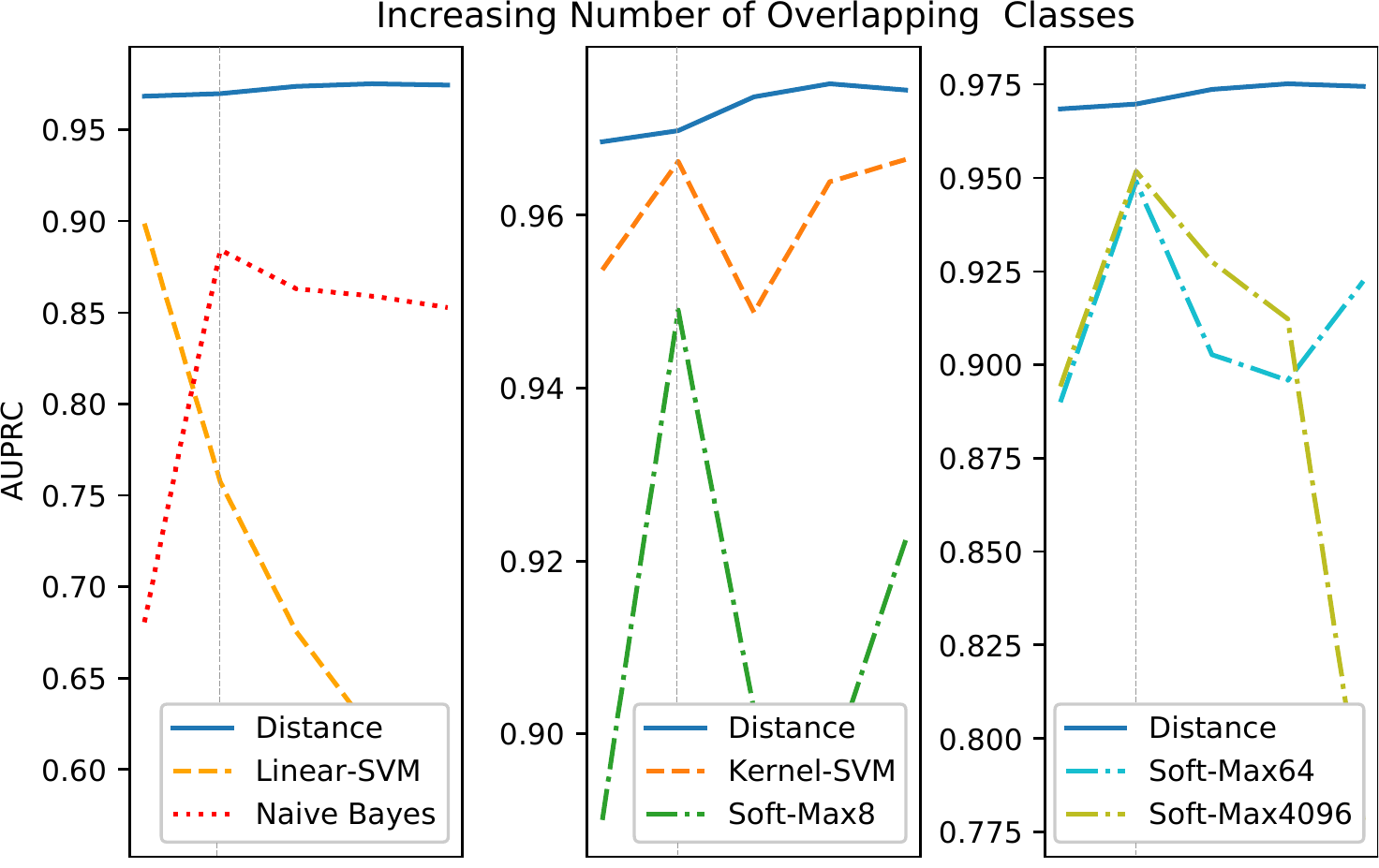}    \\
 \end{tabular}
    
\caption{ AUPRC  as unknown label overlaps increase; soft-max-X refers to training with batch-size X. 
Traditional classifiers   behave  erratically when label overlaps  are unknown;
in contrast,  Distance Classifiers are agnostic to  overlaps. 
   \label{fig:multi-label}}
\end{figure*}

 Label centric validation  seeks to estimate the likelihood   of a   feature belonging
 to a given  semantic.  If the likelihood estimate is good, there will be  a threshold (discoverable from training data) for
 determining label membership.  Rather than evaluating  algorithms at  specific thresholds, the convention is to evaluate  whether  a good threshold  is possible. One such evaluation metric is AUPRC  (area under the precision-recall curve). 
 A high AUPRC implies the existence of   a threshold where   precision
 and recall are both high. The maximum (and best) AUPRC is 1. 

 Algorithms are evaluated   on the same datasets used in \sref{sec:small_mult},
  AUPRC  scores are reported in \tref{tab:auprc}. 
 Distance Classification   is represented by two algorithms: Distance (VS) is a
 distance classifier  trained in a strictly incremental  manner;
  label likelihoods are estimated using   \eref{eq:final_veri_}.
 Distance (VW)  only  weakly  satisfies  the  incremental learning requirements,
   label likelihoods are estimated using   \eref{eq:final_veri}.
 Both Distance (VS) and  Distance (VW)    top  their respective
 categories, with   AUPRCs approaching that of discriminative classifiers. 
 
 The AUPRC metric used in  \tref{tab:auprc} is much stricter than the traditional 
 AUROC used to evaluate likelihood estimates. If measured in terms of AUROC scores,
 Distance (VW)'s  performance on MNIST and Fashion-MNIST are 0.995 and 0.963 respectively. 
 This contrasts with its AUPRC scores of $0.960$ and $0.770$ respectively.
 \footnote{ Our  evaluation  focuses only on the top one-class learning
 algorithms for these datasets. Comparisons with other one-class learning algorithms can be found in~\cite{lin21}. 
}  \newline

\noindent \textit{Novelty Detection:}
 Classifiers will eventually encounter novel instances from semantics outside of their training data;
 in such cases, it is helpful if novel instances could be identified by their 
 consistently low likelihood score across all training labels.
  To  evaluate classifiers on this metric,
   test data  is augmented with an equal number of images, randomly sampled from Flickr11k~\cite{kuo2011unsupervised}.
 An algorithm's ability to  identify novel Flickr11k images is quantified using an AUPRC score, that is recorded in
 \tref{tab:novel}. 
 The experiments show distance classifiers are surprisingly good   at  identifying  novel instances,
 a task that is difficult for even    NAT~\cite{lu2021neural},  one of the leading neural networks for  STL-10. 

 Distance classification's strength at novelty detection may be due to 
 its unique approach to learning. Unlike other top performing classifiers,
 distance classifiers are not trained to minimize a classification loss. This may make
 them  less prone to over-fitting and allow for  better generalization on    unseen classes.
 \newline
 

%

\noindent \textit{Unknown Semantic Overlap:} Another long-standing classification problem is  
 semantic overlap. If  semantic overlaps are not identified prior to training, traditional  classifiers can be ill-conditioned,
 leading to  erratic performance. To investigate this problem,  
we design the  following experiment:
\begin{itemize}
\item A set of $N$, non-overlapping  labels are chosen. 
\item New  labels are created by combining pairs of  original labels; there are  $\tensor[^{N}]{\mathrm{C}}{}_2$ new labels.
\item Algorithms are evaluated on their ability to learn  progressively larger training sets.
The first training set is  a  subset of the original labels;  training sets are progressively  expanded until they encompass   all the  original labels, this point is marked with a vertical dotted line in \fref{fig:multi-label}.
The    newly created, overlapping  labels,  are then added to the training set.
\item An algorithm's AUPRC scores on each training set is  plotted in \fref{fig:multi-label}.
\end{itemize}
 \Fref{fig:multi-label} shows that unlike traditional classifiers, 
 distance classifiers are  agnostic to semantic overlaps. This  stems
 from distance classifier's ability to represent   each semantic independently.  
 As semantics are not defined by their relation to each other, the presence of absence of semantic overlap is no longer cause for concern.

%


%

\section{Conclusion}

This paper  provides a statistical framework which reduces  the problem of  image
classification   to a modified nearest-neighbor algorithm that we term the distance
classification. Distance classifiers
are sufficiently  accurate  to act as a viable alternative for to 
a (trained) neural network's soft-max layer.
 This increases the generalizability of  a pre-trained  network as it can be: 
 incrementally updated rapidly and accurately;
accommodate huge number of classes; and posses the ability
to recognize anomalies  events.~\footnote{Code is available at: https://www.kind-of-works.com/}
\newline

\noindent\textbf{Acknowledgements:} We would like to thank Ng Hongwei of Blackmagic Design for many hours of fruitful discussions;
and the Lee Kong Chian foundation for supporting our work.

{\small
\bibliographystyle{ieee_fullname}
\bibliography{egbib}
}

\section{Appendix}

The main body focuses on  developing the best possible distance classifier.
In the appendix, we relate our proposed solution  to some popular nearest-neighbor alternatives.

\subsection{Nearest-Neighbor Classification}
\label{sec:nn_class}

\begin{figure}[tp]
\centering
\includegraphics[width=1\linewidth]{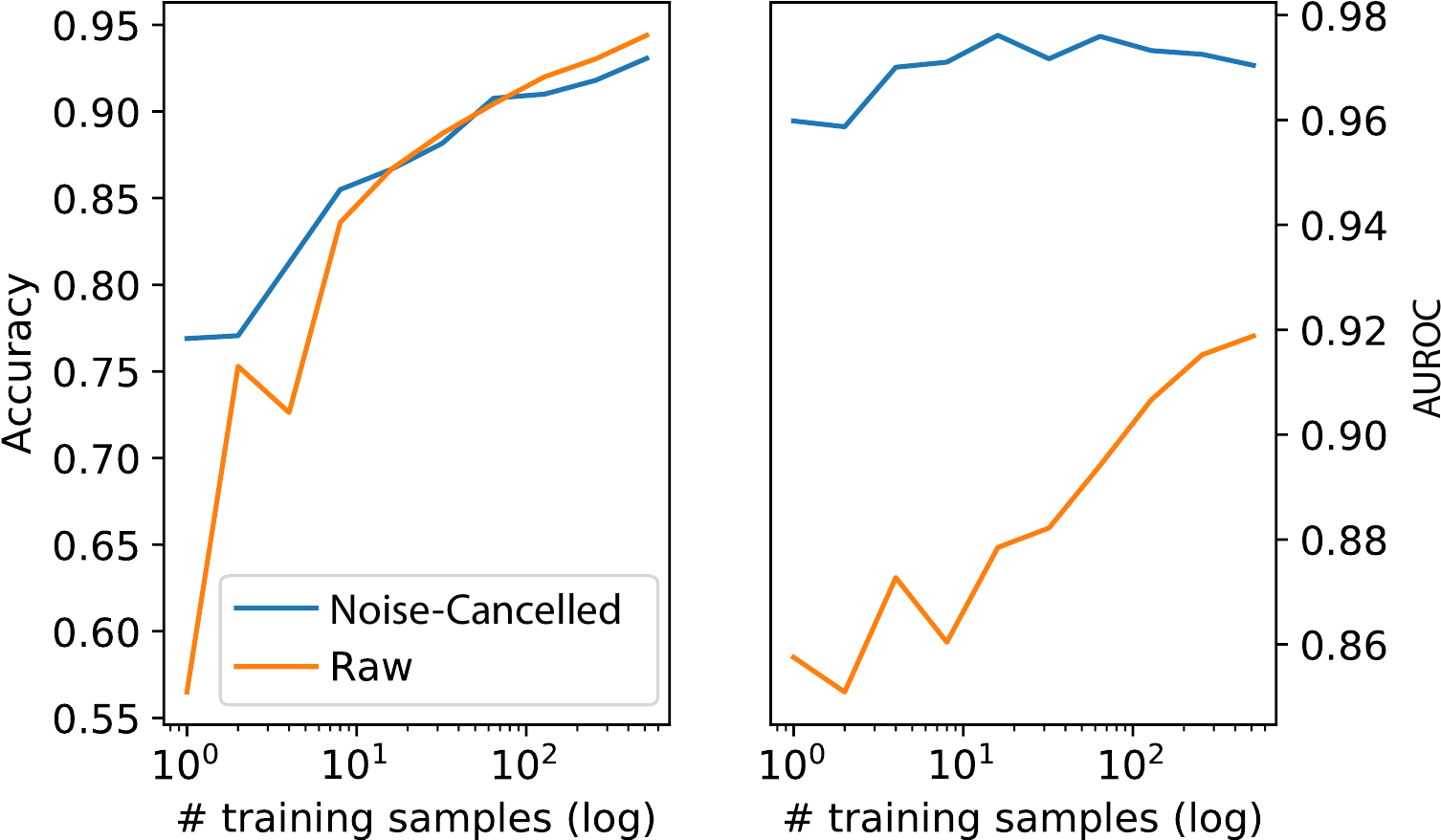}
\caption{Nearest-neighbor classification on STL-10~\cite{coates2011analysis} with raw and reference based noise cancelled distances. Noise cancelled distances provide significantly better validation, as shown by the AUROC score. 
   \label{fig:nn}}
\end{figure}

Finally, no discussion of distance based classification can be complete without mentioning the
nearest-neighbor classifier. This section introduces reference based
noise cancelled nearest-neighbor classifier. \Fref{fig:nn} shows that (similar to  our other distance
based classifier),  noise cancellation greatly improves the nearest neighbor classifier's  
validation capability. 
The  derivation is provided below.

Let $\mathbf{E}$ be the ideal, common  generator of all instances in  a dataset and
$\bm$ be its mean. As $\bm$  is  constant relative to the process
of generating individual instances, from \eref{eq:base1} the distance of all ideal instances
to $\bm$ will be a constant, which we denote as $c_\bm$. \ie, if  $\bx_t, \, \bx_{t'}$ are two ideal instances
\begin{equation}
\label{eq:nn_cm}
a.s. \quad \| \bx_t - \bm \| = \| \bx_{t'} - \bm \| = c_\bm.
\end{equation}

From \eref{eq:dream_mean},  the distance of the noisy features $\bx(t), \, \bx(t')$ from $\bm$ is:
\begin{equation}
\begin{split}
\label{eq:nn_cm_n}
a.s. \quad & \|\bx(t) - \bm\|^2  \\
\approx &\|\bx_t - \bm\|^2+ \|\bn (t)\|^2 = c_\bm +\|\bn (t)\|^2 , \\
a.s. \quad & \|\bx(t') - \bm\|^2 \\
\approx & \|\bx_{t'} - \bm\|^2  + \|\bn (t')\|^2 =  c_\bm +\|\bn (t')\|^2.
\end{split}
\end{equation}
From \eref{eq:dream_feat}, the distance of $\bx(t)$ from $\bx(t')$ is
\begin{equation}
\label{eq:nn_copy}
 \|\bx(t) - \bx(t')\|^2 \approx \|\bx_t - \bx_{t'} \|^2 + \|\bn (t)\|^2 + \|\bn (t')\|^2.
\end{equation}
Thus, combining  \eref{eq:nn_cm_n} and \eref{eq:nn_copy}, a
noise cancelled nearest-neighbor   distance be can be defined as:
\begin{equation}
\label{eq:noise_can_neigh}
\begin{split}
	& f^2(\bx(t),  \bx(t'), \bm)  \\
=  & \|\bx(t) - \bx(t') \|^2 - \| \bx(t) - \bm \|^2  - \| \bx(t') - \bm \|^2 \\
\approx  & \|\bx_t - \bx_{t'}\|^2 - 2c_\bm, \\
\end{split}
\end{equation}
where $f^2(\bx(t),  \bx(t'), \bm)$  approximates the ideal  squared  distance with a constant offset, $ -2c_\bm$.

In terms of  practical effectiveness, reference based noise canceled nearest-neighbor, is similar  to that of  standard normalization. 
However, the ability to achieve a normalization like effect with a different algorithm,
helps validate our theory.  It also provides a chance to reconsider,    the classic   algorithm of undergraduate textbooks,  from a new perspective.

\subsection{Distance vs  Nearest-Neighbor Classification}
\label{sec:compare_nn}

Finally, it would be instructive to compare our distance classifier with a traditional  nearest   
neighbor classification algorithm. For this task, we consider both euclidean
and cosine distance classifiers with and without normalization / centering. 
They against our distance classifier in \tref{tab:cos_dist}, which reports both classification
accuracy and  validation AUPRC.

As predicted, all the nearest-neighbor classification accuracies are respectable, with only minor variations
across algorithms. However,  there are large differences in AUPRC. For Euclidean nearest-neighbor,
normalization significantly improves  AUPRC, which shell theory and \sref{sec:norm} explain in terms of a noise cancellation effect.
A similar improvement occurs when applying the cosine distance to centered data-points. However, we 
do not offer an analytical explanation  because cosine distances are not translational invariant; and thus, 
cannot be trivially analyzed using shell theory. 

Although normalization significantly improves validation AUPRC of traditional nearest-neighbor algorithms, the scores
remain significantly below that of our distance classifier. If normalization cannot be employed (such as in the context
of strictly incremental learning), our distance classifier will have significantly high validation AUPRC than either of the traditional
nearest-neighbor techniques.

\begin{table*}[htp]
\footnotesize
\centering
\begin{tabular}{l|l|c|c|c|c|c|c}
\toprule
&\multicolumn{6}{c}{\textit{Classification Accuracy }}\\
\hline
\makecell[l]{Strictly \\ Incremental} & & MNIST &\makecell{\small Fashion-MNIST}  &\makecell{STL-10}& \makecell{Internet STL-10} & \makecell{SUN-Small} & ASSIRA  \\
\hline
\multirow{2}{*}{\makecell[l]{  No }}  
& Cos. NN (raw)  & 0.968 & \textbf{0.852} & 0.952 & 0.887 &0.906 & 0.982\\
& Euc. NN (raw) & \textbf{0.969}& 0.850 & 0.955 & 0.865 & 0.885 & \textbf{0.986}\\
& Dist. Classifier (raw) & {0.955} & {0.825} & \textbf{0.970} & \textbf{0.907} & \textbf{0.929} & \textbf{0.986}\\
\hline
	\multirow{2}{*}{\makecell[l]{ Yes }}    
& Cos. NN (centered) & \textbf{0.969} & \textbf{0.858} & 0.955 & 0.888 &0.903 & 0.979\\
& Euc. NN (normalized) & \textbf{0.969} & 0.858 & 0.955 & 0.888 &0.898 & 0.979\\
& Dist. Classifier (raw)  & {0.955} & {0.825} & \textbf{0.970} & \textbf{0.907} & \textbf{0.929} & \textbf{0.986}\\
\hline
&\multicolumn{6}{c}{\textit{Validation AUPRC  }}\\
\hline
 \makecell[l]{Strictly \\ Incremental}& & MNIST &\makecell{\small Fashion-MNIST}  &\makecell{STL-10}& \makecell{Internet STL-10} & \makecell{SUN-Small} & ASSIRA \\
\hline
\multirow{2}{*}{\makecell[l]{  No  }}  
& Cos. NN (raw) & 0.767 & 0.624 & 0.912 & 0.851 & 0.764 & 0.976\\
& Euc. NN (raw) & 0.784 & 0.673& 0.764 & 0.589 & 0.562 & 0.915\\
&Dist. Classifier (VS) & \textbf{0.980} &    \textbf{0.852}       & \textbf{0.984} & \textbf{0.931} & \textbf{0.941} & \textbf{0.985}\\
\hline
	\multirow{2}{*}{\makecell[l]{  Yes }}    
& Cos. NN (centered) & 0.923 & 0.667 & 0.915 & 0.855 & 0.825 & 0.958\\
& Euc. NN (normalized)  & 0.923 & 0.667& 0.922 & 0.855 &0.805 &0.958\\
& Dist. Classifier (VW) & \textbf{0.960} & \textbf{0.770} & \textbf{0.984} & \textbf{0.951}  & \textbf{0.969} & \textbf{0.999}\\
\hline

\end{tabular}
\caption{Comparing   distance classification with traditional nearest-neighbors. Accuracy 
scores are similar; however, distance classification has  higher validation AUPRC. This is especially notable
in the full incremental case, where the classifier must be trained without knowledge of the true dataset mean. 
\label{tab:cos_dist}}
\end{table*}

\end{document}